\documentclass[10pt,journal,letterpaper,compsoc,twoside]{IEEEtran}
\usepackage[margin=0.5in]{geometry}

\usepackage[english]{babel}
\usepackage{array}
\usepackage{amsmath,amssymb,bbm}

\usepackage{graphicx}
\usepackage{epsf, epsfig}
\usepackage{epstopdf}
\usepackage{enumitem}

\usepackage[numbers]{natbib}
\usepackage{hyperref}

\usepackage{booktabs,xcolor,siunitx}
\definecolor{lightgray}{gray}{0.9}
\usepackage{algorithm}
\usepackage{algorithmic}

\makeatletter
\newcommand\fs@norules{\def\@fs@cfont{\bfseries}\let\@fs@capt\floatc@ruled
  \def\@fs@pre{}%
  \def\@fs@post{}%
  \def\@fs@mid{\kern3pt}%
  \let\@fs@iftopcapt\iftrue}
\makeatother
\floatstyle{norules}
\restylefloat{algorithm}

\catcode`\>=\active \def>{
\fontencoding{T1}\selectfont\symbol{62}\fontencoding{\encodingdefault}}
\catcode`\|=\active \def|{
\fontencoding{T1}\selectfont\symbol{124}\fontencoding{\encodingdefault}}
\newcommand{\assign}{:=}
\newcommand{\nobracket}{}
\newcommand{\nocomma}{}
\newcommand{\nosymbol}{}
\newcommand{\tmop}[1]{\ensuremath{\operatorname{#1}}}
\newcommand{\tmtextbf}[1]{{\bfseries{#1}}}
\newenvironment{itemizedot}{\begin{itemize} }{\end{itemize}}

\begin{document}
\title{Probabilistic Multigraph Modeling for Improving\break 
the Quality of Crowdsourced Affective Data}
\author{Jianbo Ye, Jia Li, Michelle G. Newman, Reginald B. Adams, Jr. and James Z. Wang\thanks{Manuscript received \ \ \ ; revised \ \ \ .}
\thanks{J. Ye and J. Z. Wang are with the College of Information Sciences and Technology, The Pennsylvania State University, University Park, PA 16802, USA. J. Li is with the Department of Statistics, The Pennsylvania State University, University Park, PA 16802, USA. M. Newman and R. B. Adams, Jr. are with the Department of Psychology, The Pennsylvania State University, University Park, PA 16802, USA.
(e-mails: \{jxy198,jiali,mgn1,radams,jwang\}@psu.edu)}
}

\markboth{IEEE TRANSACTIONS ON Affective Computing,,~Vol.~1, No.~1,~January~2016}
{Ye \MakeLowercase{\textit{et al.}}: Probabilistic Multigraph Modeling ...}

\maketitle

\begin{abstract}
We proposed a probabilistic approach to joint modeling of
participants' \textit{reliability} and humans' \textit{regularity}
in crowdsourced affective studies. 
{Reliability} measures how likely 
a subject will respond to a question {seriously};
and {regularity} measures how often 
a human will agree with other seriously-entered responses coming from a 
targeted population. Crowdsourcing-based studies or experiments,  
which rely on human self-reported affect, pose additional challenges
as compared with 
typical crowdsourcing studies that attempt to acquire {\it concrete non-affective} labels of objects. 
The reliability of participants has been 
massively pursued for typical non-affective crowdsourcing studies, 
whereas the regularity of humans in an affective experiment 
in its own right has not been thoroughly considered.  
It has been often observed that different individuals exhibit different feelings on 
the same test question, which does not have a sole correct response in the first place.
High reliability of responses from one individual thus cannot conclusively result in
high consensus across individuals. Instead, globally testing consensus
of a population is of interest to investigators.
Built upon the agreement multigraph among tasks and workers, our probabilistic model differentiates 
subject regularity from population reliability. 
We demonstrate the method's effectiveness for in-depth robust analysis of 
large-scale crowdsourced affective data, including emotion and aesthetic assessments collected 
by presenting visual stimuli to human subjects. 
\end{abstract}

\begin{IEEEkeywords}
Emotions, human subjects, crowdsourcing, probabilistic graphical model, visual stimuli
\end{IEEEkeywords}

\IEEEpeerreviewmaketitle

\section{Introduction}
Humans' sensitivity to affective stimuli intrinsically varies from one person
to another. Differences in gender, age, society, culture, personality, 
social status, and personal experience can contribute to its 
high variability between people. Further, inconsistencies 
may also exist for the same individual across environmental contexts and current mood or affective state. 
The {causal} effects and factors for such affective experiences 
have been extensively investigated, as evident in the literature on 
psychological and human studies, where controlled 
experiments are commonly conducted within a small group
of human subjects --- to ensure the reliability of collected data. 
To complement the shortcomings of those 
controlled experiments, ecological psychology 
aims to understand how objects and things 
in our surrounding environments effect human behaviors and affective experiences,
in which {\it real-world} studies are favored over those within
artificial laboratory environments~\citep{barker1968ecological,gibson1966senses}.
The key ingredient of those ecological approaches is the availability of
large-scale data collected from human subjects, 
remedying the high complexity and heterogeneity that 
the real-world has to offer. With the growing attention
on affective computing (initiated from the seminal discussion \citep{picard1997affective} to 
recent communications~\citep{marsella2014computationally}),
multiple data-driven approaches have been developed to understand
what particular environmental factors drive the feelings of humans~\citep{datta2006studying,lu2012shape},
and how those effects differ among various sociological 
structures and between human groups.

One crucial hurdle for those affective computing approaches is the lack
of full-spectrum annotated stimuli data at a large scale. 
To address this bottleneck, crowdsourcing-based approaches are highly helpful
for collecting uncontrolled human data from anonymous participants~\cite{howe2006rise}.
In a recent study reported in~\cite{xin2016}, 
anonymous subjects from the Internet were recruited to annotate
a set of visual stimuli (images): 
at each time point, after being presented with an image stimulus, participants were asked 
to assess their personal psychological experiences 
using ordinal scales for each of the affective dimensions: valence, arousal, dominance and likeness (which means the degree of appreciation in our context). This study also collected
demographics data to analyze individual difference predictors of 
affective responses. 
Because labeling a large number of visual stimuli can become tedious, 
even with crowdsourcing, each image stimulus was examined by only a few subjects.
This study allowed tens of thousands of images to obtain at least
one label from a participant, which created a large data set for 
environmental psychology and automated emotion analysis of images. 

One interesting question to investigate, however, is {\it whether the affective labels provided by subjects are reliable}. {A related question is how to separate spammers from reliable subjects, or at least to narrow the scope of data to a highly reliable subgroup. Here, spammers are defined as those participants who provide answers without 
serious consideration of the presented questions. No answer from a statistical perspective is known yet for crowdsourced affective data. }

A great difficulty in analyzing affective data is caused by the absence of ground truth in the first place, that is, there is no {\it correct} answer for evoked emotion. It is generally accepted that even the most reliable subjects can naturally have varied emotions. Indeed, with variability among human responses anticipated, psychological studies often care about questions such as {where humans are emotionally consistent and where they are not}, and {which subgroups of humans are more consistent} than another.  Given a population, many, if not the vast majority of stimuli may not have a consensus emotion at all.
{Majority voting or (weighted) averaging to force an "objective truth" of the emotional response or probably for the sake of convenience, as is routinely done in affective computing so that classification on a single quantity can be carried out, is a crude treatment bound to erase or disregard information essential for 
many interesting psychological studies, {\it e.g.},
to discover connections between varied affective responses and varied demographics.}

\begin{figure}
\parbox{2.6cm}{
\includegraphics[width=2.5cm]{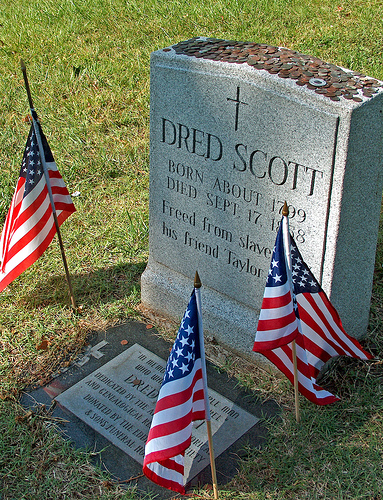}
}\hfill
\parbox{6cm}{
\begin{tabular}{ccc} 
Annotator ID & Valence & Reliability \\ \hline
3474 & 5.1/8 & 0.08 \\
2500 & 0.0/8 & 0.56 \\
3475 & 0.0/8 & 0.34\\
2540 & 8.0/8 & 0.04 \\
\end{tabular}\\

\quad Image Confidence: 75\% ($\le $ 90\%)
}
\caption{An example illustrating one may need to acquire more reliable labels,
ensuring the image confidence is more than 0.9.}\label{fig:oneimage}
\end{figure}

{The involvement of spammers as participating subjects introduces an extra source of variation to the emotional responses, which unfortunately is tangled with the "appropriate" variation.} If responses associated with an image stimulus contain answers by spammers, the inter-annotator
variation for the specific question could be as large as the variation across different questions, reducing the robustness of any analysis. An example is shown in Fig.~\ref{fig:oneimage}. Most annotators labeling this image are deemed unreliable, and two of them are highly susceptible as spammers according to our model. 
Investigators may be recommended to eliminate this image or acquire more reliable labels for its use. {Yet, one should not be swayed by this example into the practice of discarding images that solicited responses of a large range. Certain images are controversial in nature and will stimulate quite different emotions to different viewers.  Our system acquired the reliability scores shown in Fig.~\ref{fig:oneimage} by examining the entire data set; the data on this image alone would not be conclusive, in fact, far from so.}

{Facing the intertwined "appropriate" and "inappropriate" variations in the subjects as well as the variations in the images, we are motivated to unravel the sources of uncertainties by taking a global approach. The judgment on the reliability of a subject cannot be a per-image decision, and has to leverage the whole data. Our model was constructed to integrate these uncertainties, attempting to discern them with the help of big data. In addition, due to the lack of ground truth labels, we model the relational data that code whether two subjects' emotion responses on an image agree, bypassing the thorny questions of what the true labels are and if they exist at all.   }


For the sake of automated emotion analysis of images, one also needs to narrow the scope to parts of data, each of which have sufficient number of qualified labels. Our work computes image confidences, which can support off-line data filtering or guide on-line budgeted crowdsourcing practices. 

\begin{figure*}[ht!]
\centering
\begin{tabular}{cccccc}
\includegraphics[height=2.8cm]{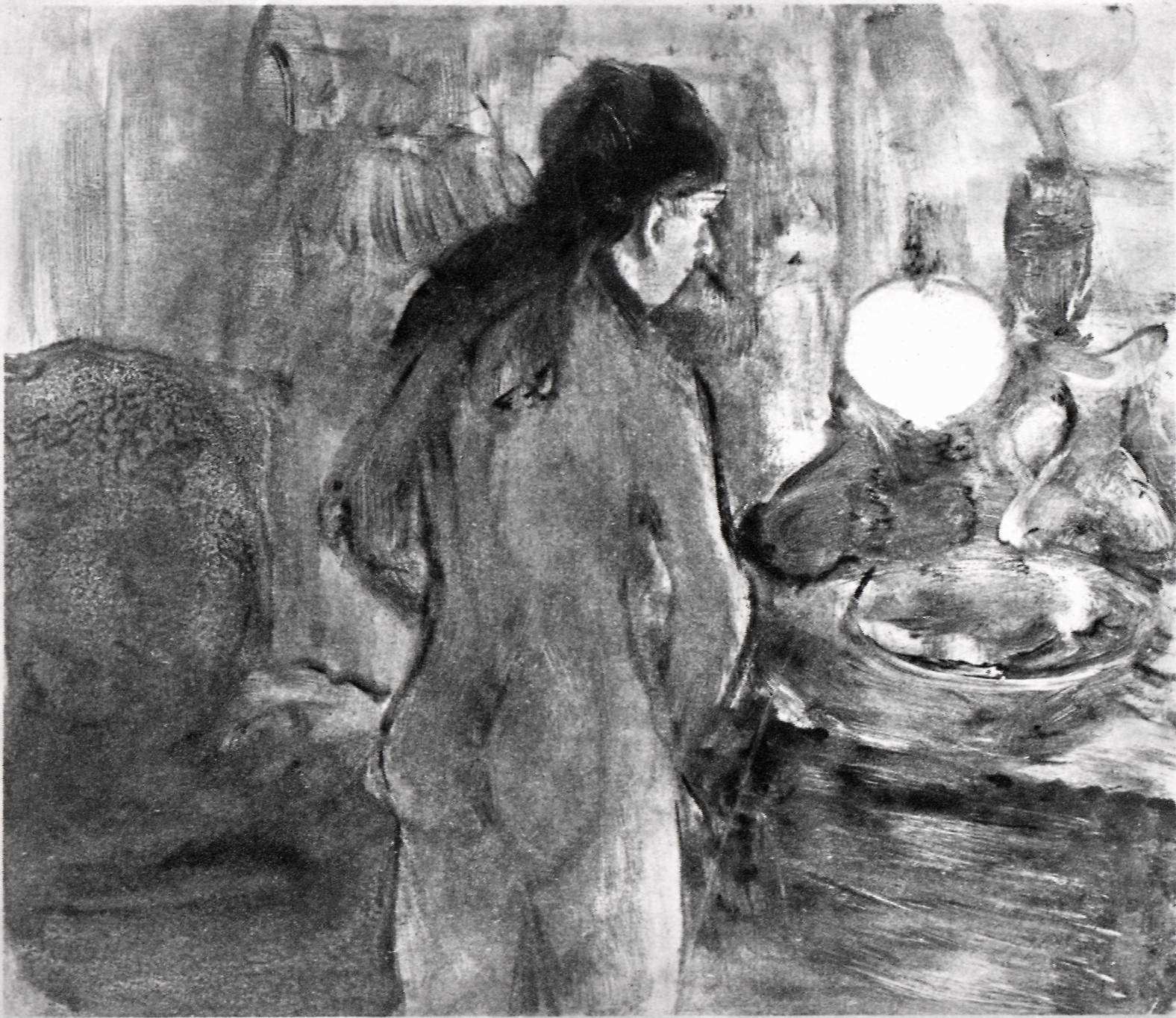} &
\includegraphics[height=2.8cm]{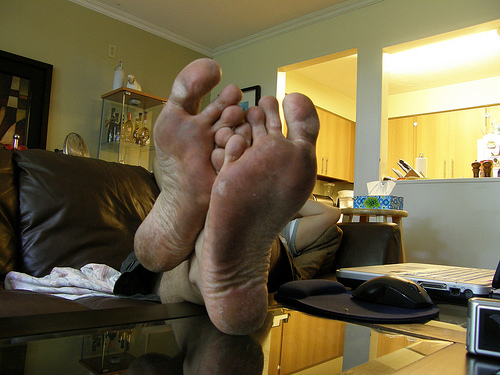} &
\includegraphics[height=2.8cm]{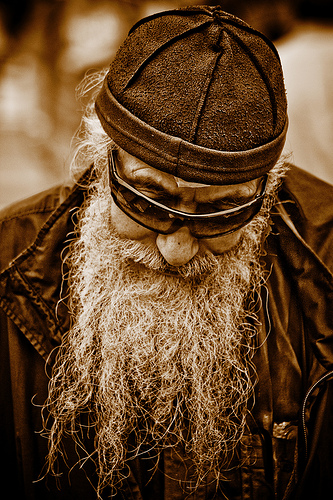} &
\includegraphics[height=2.8cm]{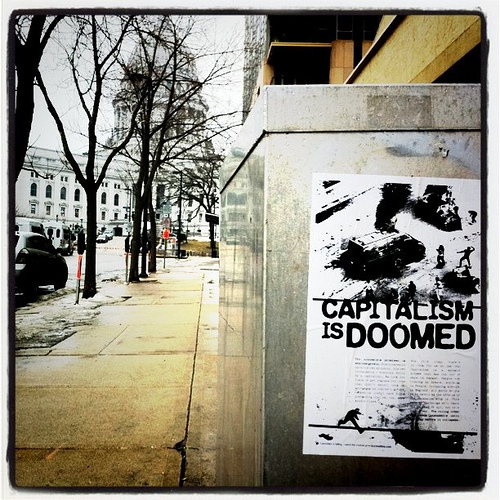} &
\includegraphics[height=2.8cm]{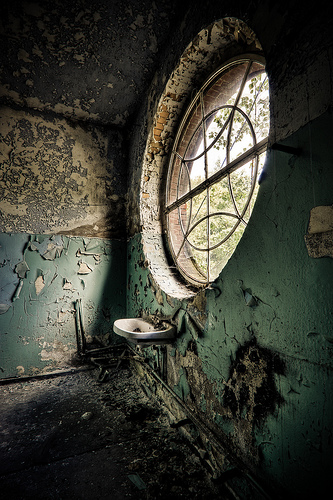}&
\includegraphics[height=3cm]{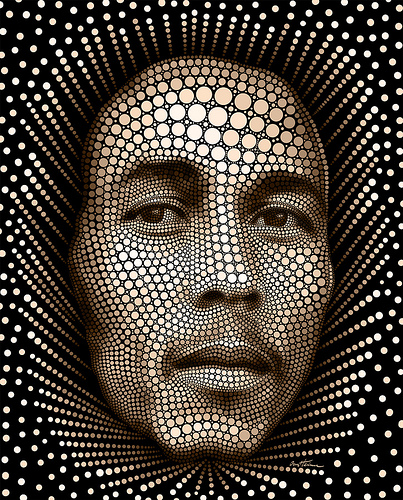}\\[-0.2cm]
\small raw avg.: 4.06 out of 8 &
4.1 $\rightarrow$ 2.94 &
4.78 $\rightarrow$ 3.33 &
4.25 $\rightarrow$ 1.9 &
4.54 $\rightarrow$ 2.75 &
4.53 $\rightarrow$ 3\\[-0.2cm]
$\rightarrow$ new: 2.51 out of 8  & &&&\\
\end{tabular}
\begin{tabular}{ccccc}
\includegraphics[height=2.7cm]{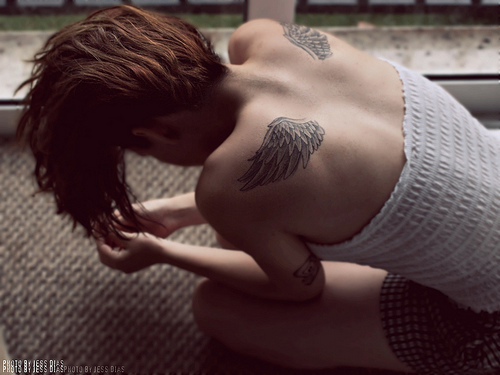} &
\includegraphics[height=2.7cm]{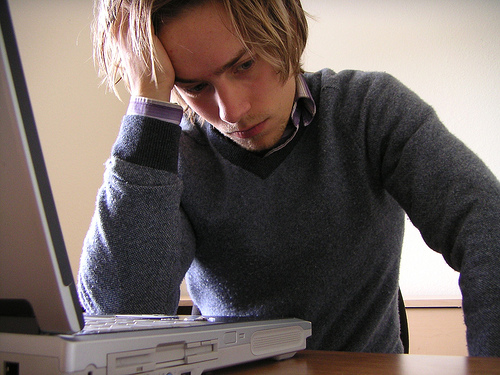} &
\includegraphics[height=2.7cm]{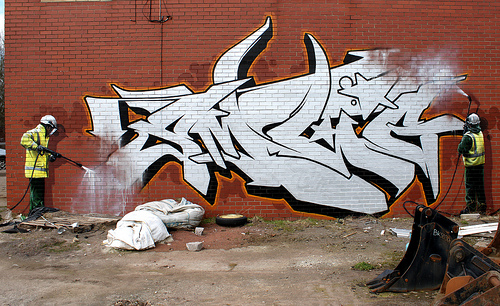} &
\includegraphics[height=2.7cm]{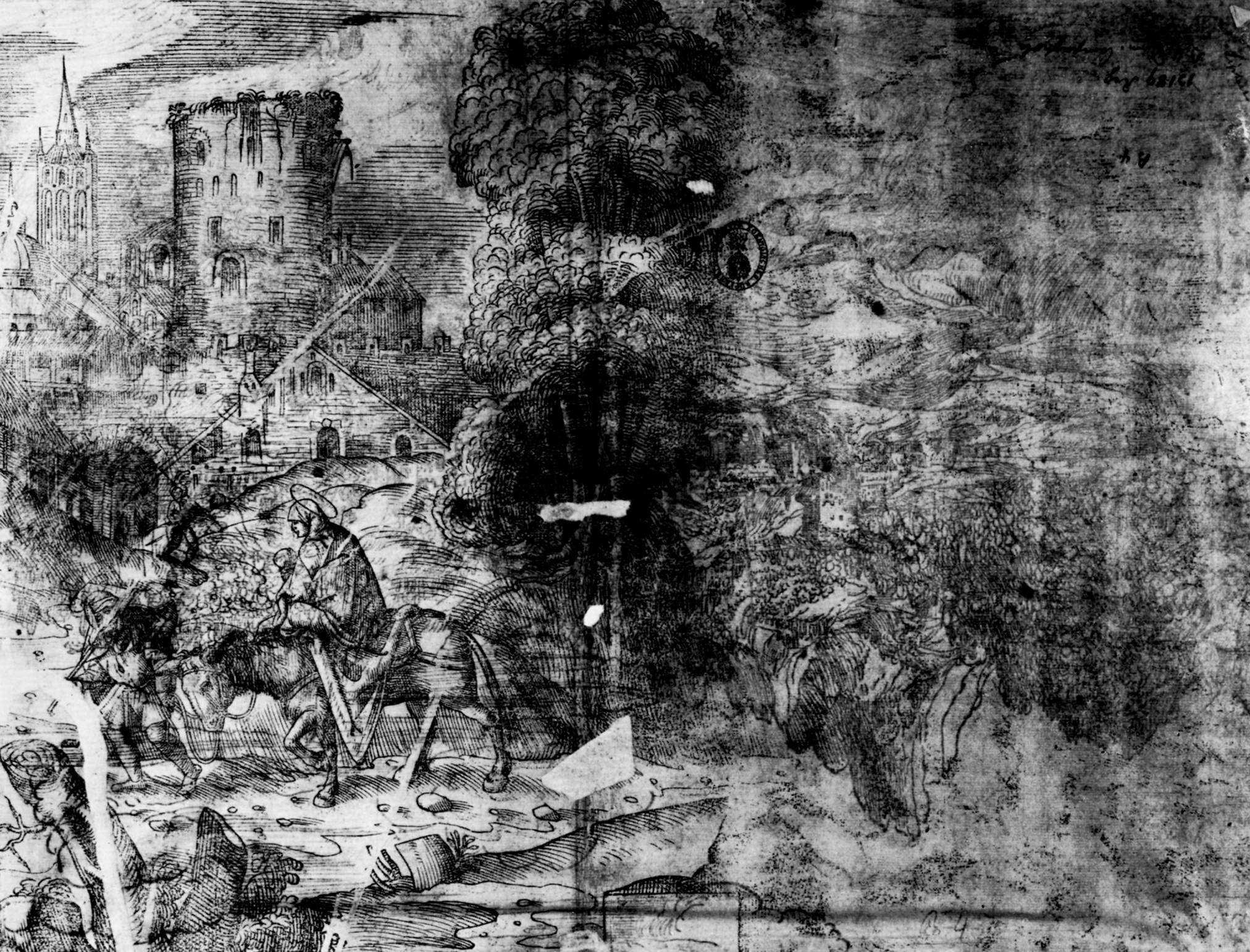} &
\includegraphics[height=2.7cm]{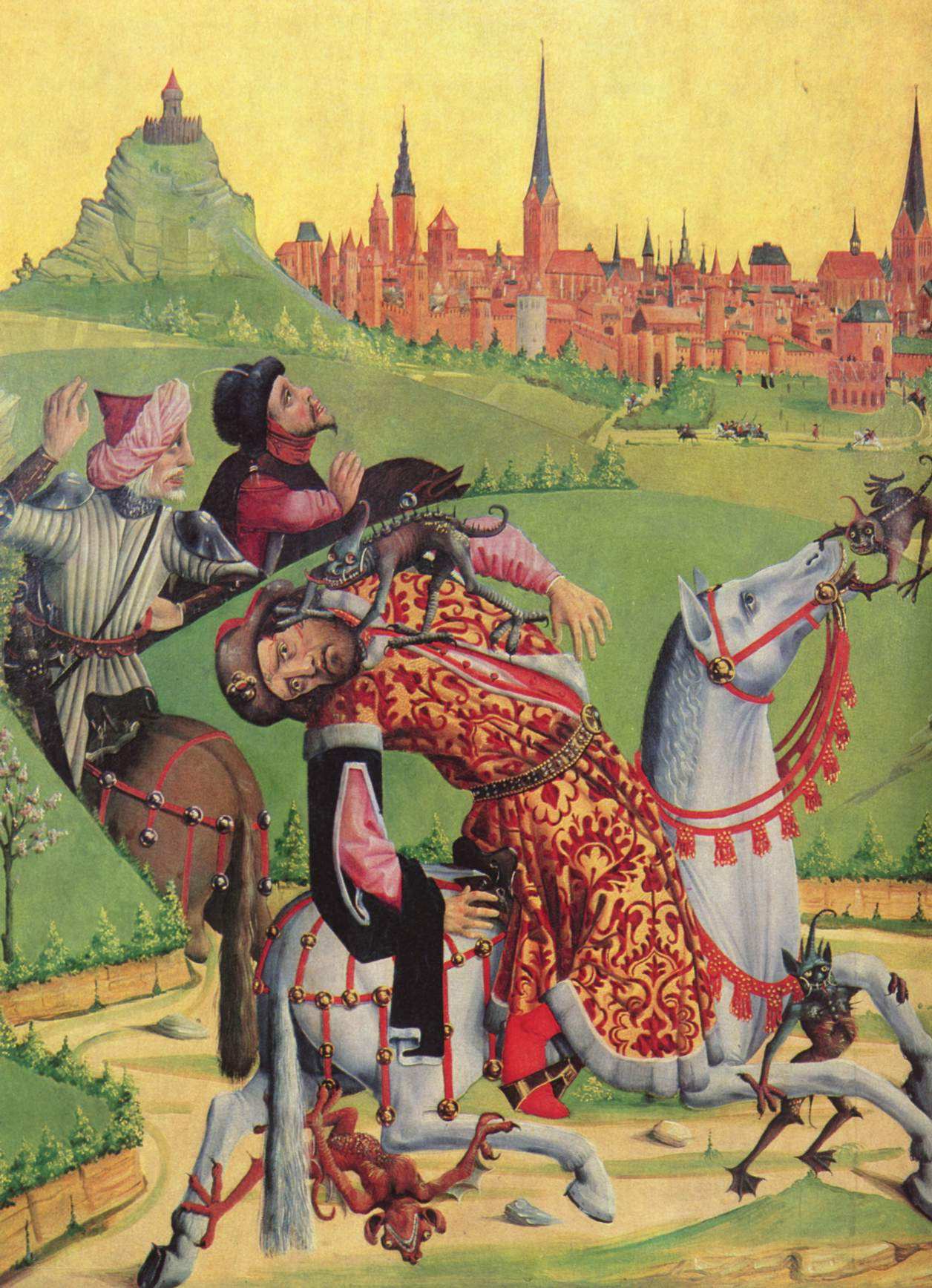}\\[-0.2cm]
\small 4.06 $\rightarrow$ 3.03 &
4.05 $\rightarrow$ 2.87 &
4.7 $\rightarrow$ 2.06 &
5.08 $\rightarrow$ 3.94 &
5.24 $\rightarrow$ 3.93
\end{tabular}
\begin{tabular}{cccccc}
\includegraphics[height=2.8cm]{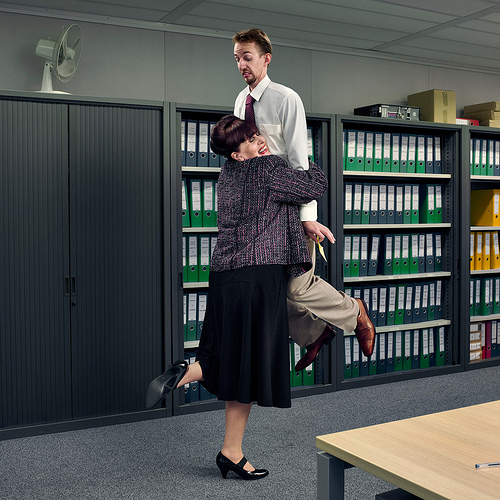} &
\includegraphics[height=2.8cm]{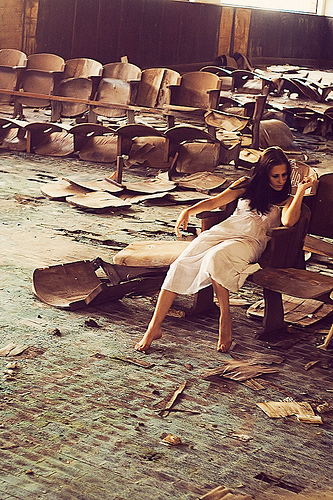} &
\includegraphics[height=2.8cm]{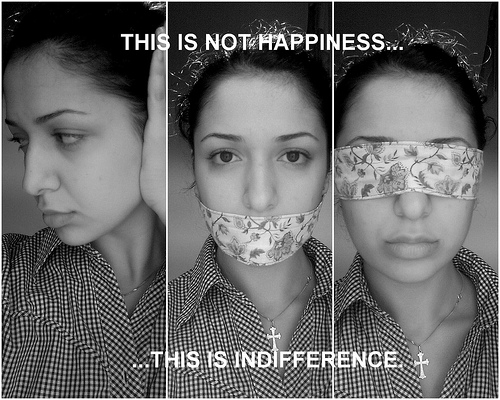} &
\includegraphics[height=2.8cm]{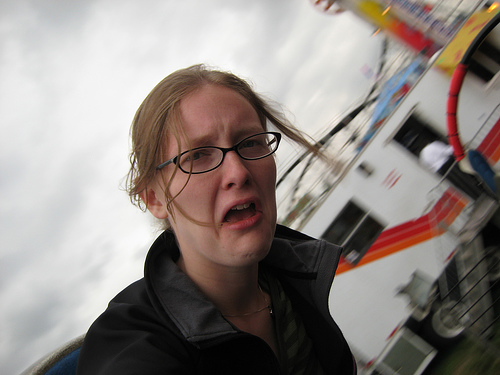} &
\includegraphics[height=2.8cm]{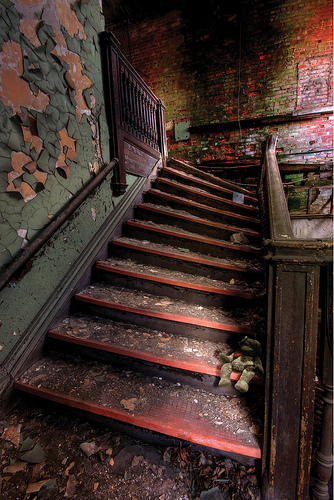} &
\includegraphics[height=2.8cm]{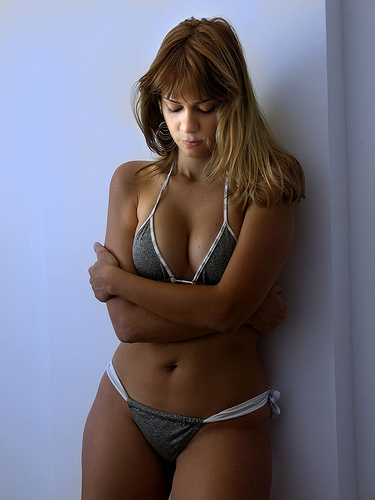}\\[-0.2cm]
\small
5.02 $\rightarrow$ 3.58 &
5.7 $\rightarrow$ 3.87 &
5.6 $\rightarrow$ 3 &
5.17 $\rightarrow$ 3.19 &
5.32 $\rightarrow$ 2.98 &
5.38 $\rightarrow$ 3.76
\end{tabular}
\caption{Images shown are considered of lower valence 
than their average valence ratings 
({\it i.e.}, evoking a higher degree of negative emotions) 
after processing the data set using our proposed method. 
Our method eliminates the contamination introduced by spammers. The range of valence ratings is between 0 and 8.}
\label{fig:example}
\end{figure*}

\begin{figure*}
\centering
\begin{tabular}{ccccc}
\includegraphics[height=3cm]{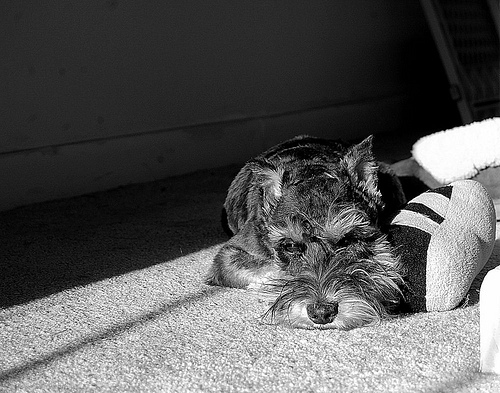} &
\includegraphics[height=3cm]{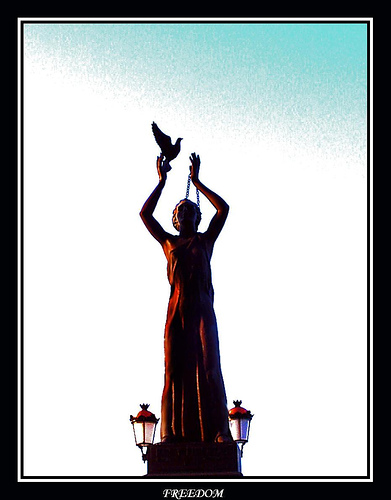} &
\includegraphics[height=3cm]{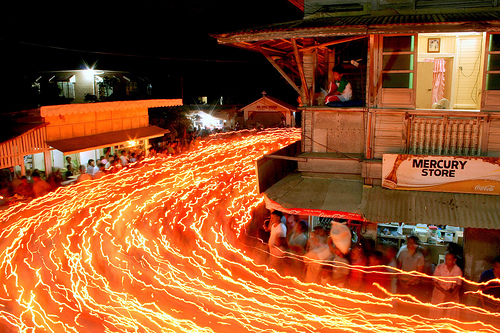} &
\includegraphics[height=3cm]{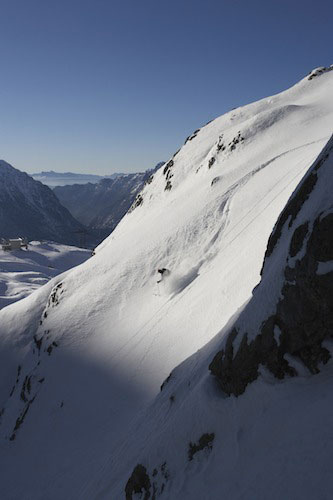} &
\includegraphics[height=3cm]{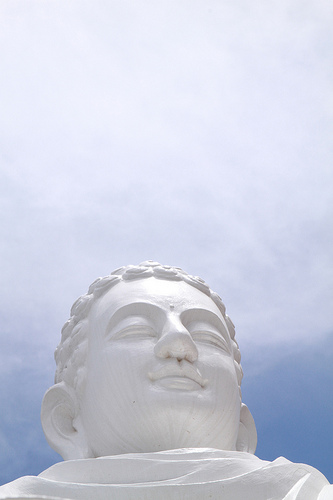}\\[-0.2cm]
2.63 $\rightarrow$ 3.77 &
2.8 $\rightarrow$ 4.14 &
3.0 $\rightarrow$ 4.7 &
4.4 $\rightarrow$ 6.21 &
4.7 $\rightarrow$ 6.26
\end{tabular}
\caption{Images shown are considered of higher valence 
than their average valence ratings ({\it i.e.}, evoking a higher degree of
positive emotions) 
after processing the data set using our proposed method. 
Our method again eliminates the contamination introduced by spammers. The range of valence ratings is between 0 and 8.}\label{fig:example2}
\end{figure*}

In summary, systematic analysis of crowdsourced 
affective data is of great importance to human subject studies 
and affective computing, while remains an open question.
To substantially address the aforementioned challenges and expand
the evidential space for psychological studies, 
we propose a probabilistic approach, called {\bf Gated
Latent Beta Allocation (GLBA)}. This method computes maximum a posteriori probability (MAP) estimates of 
each subject's reliability and regularity based on a variational expectation-maximization (EM) framework. 
With this method, investigators running affective human subject studies can substantially reduce or eliminate
the contamination caused by spammers, hence improve the quality and usefulness of collected data (Fig.~\ref{fig:example}).

\subsection{Related Work}
Estimating the reliability of subjects is necessary in 
crowdsourcing-based data collection because the incentives of participants and the interest of researchers diverge. 
There were two levels of assumptions explored for the crowdsourced data, 
which we name as the first-order
assumption (A1) and the second-order assumption (A2). Let a task be the provision of emotion responses for one image. Consider a task or test conducted by a number of participants. Their responses within this task form a subgroup of data. 
\begin{description}
\item [A1] There exists a true label of practical interest for each task.
 The dependencies between collected labels are mediated by this unobserved true label, 
 of which noisy labels are otherwise conditionally independent. 
\item [A2] The uncertainty model for a subgroup of data
does not depend on its actual specified task. 
The performance of a participant is consistent across subgroups of data subject to a single fixed effect.
\end{description}

Existing approaches that model the complexities of tasks or reliability of participants often require
 one or both of these two assumptions. Under the umbrella of assumption A1, most probabilistic 
 approaches using the observer models  ~\citep{dawid1979maximum,hui1980estimating,smyth1995inferring,demartini2012zencrowd}
 focus on estimating the ground truth from multiple noisy labels. For example,
 the modeling of one reliability parameter per subject is an established practice for estimating 
 the ground truth label~\cite{demartini2012zencrowd}. For the case of categorical
 labels, modeling of one free parameter per class per subject is a more general 
 approach~\cite{dawid1979maximum,raykar2010learning}.
 Our approach does not model the ground truth of labels, hence it is not viable to compare our approach with other methods in this regard. Instead, we sidestep this issue to tackle whether the labels from one subject can agree 
 with labels from another on a single task. 
 Agreement is judged subject to a preselected criterion.
Such treatment may be more realistic as a means to process sparse ordinal labels for each task. 

Assumption A2 is also widely exploited among methods, often conditioned on A1.
It assumes that all of the tasks have the same level of difficulty~\citep{liu2012variational,raykar2012eliminating}.
Modeling one difficulty parameter per task 
has been explored in~\cite{whitehill2009whose} for categorical labels. 
However, in our approach, task difficulty is modeled as a random effect 
without subscribing a task-specific parameter. 
Wisely choosing the modeling complexity and assumptions should 
be {based} on availability and purity of data.  
As suggested in~\cite{sheshadri2013square}, 
more complexity in a model could challenge the statistical estimation subject 
to the constraint of real data. Choices with respect to our model attempted to
properly analyze the affective data we obtained. 

If the mutual agreement rate between two participants
 does not depend on the actual specified task ({\it i.e.}, when A2 holds), we can essentially convert the 
 resulting problem to a graph mining problem, where subjects are vertices, agreements are edges, and the
 proximity between subjects is modeled by how likely they agree with each other in a general sense. 
 Probabilistic models for such relational data
 can be traced back to early stochastic blockmodels~\citep{wang1987stochastic,nowicki2001estimation},
 latent space model~\citep{hoff2002latent},
 and their later extensions with mixed membership~\citep{airoldi2009mixed,kim2012latent} and
 nonparametric Bayes~\citep{kemp2006learning}. 
 We adopt the idea of mixed memberships wherein two particular modes of memberships
 are modeled for each subject, one being the reliable mode and the other the random mode. 
For the random mode, the behavior is assumed to be shared across different subjects, whereas the 
 regular behaviors of subjects in the reliable mode are assumed to be different. 
 Therefore, we can extend this framework from graph to multigraph in the
 interest of crowdsourced data analysis. Specifically,
 data are collected as subgroups, each of which is composed of 
 a small agreement graphs for a single task, such that the covariate within a subgroup is modeled. 
Our {approach} does not rely on A2.
Instead, it models the random effects added to subjects' performance in each task via the multigraph approach.
{Assumption A1 and A2 implies a bipartite graph structure between tasks and subjects.
In contrast, our approach starts from the multigraph structure among subjects that is coordinated by tasks.
Finding the proper and flexible structure that data possess is crucial for modeling~\cite{kemp2008discovery}.} 
 

\subsection{Our Contributions}
 To our knowledge, this is the first attempt to connect probabilistic observer models with probabilistic graphs, and to explore modeling at this complexity from the joint perspective. 
 We summarize our contributions as follows:
 \begin{itemize}
\item We developed a probabilistic multigraph model to analyze crowdsourced data and its
approximate variational EM algorithm for estimation. {The new method, accepting the intrinsic variation in subjective responses, does not assume the existence of ground truth labels, in stark contrast to previous work having devoted much effort to obtain objective true labels.}
\item {
Our method exploits the relational data in the construction and application of the statistical model. Specifically, instead of the direct labels, the pair-wise status of agreement between labels given by different subjects is used. As a result, the multigraph agreement model is naturally applicable to more flexible types of responses, easily going beyond binary and categorical labels. Our work serves as a proof of concept for this new relational perspective.}
\item Our experiments have validated the effectiveness of our approach on real-world affective data. Because our experimental setup was of a larger scale and more challenging {than settings  addressed by existing methods, we believe our method can fill some gaps for demands in the practical world, for instance, when gold standards are not available.}
\end{itemize}

\section{The Method}
In this section, we describe our proposed method. Let us present the mathematical notations first. A symbol with subscript omitted always
indicates an array, {\it e.g.}, $x=(\ldots,x_i,\ldots)$. The 
arithmetic operations perform over arrays in
the element-wise manner, {\it e.g.}, $x+y=(\ldots,x_i+y_i,\ldots)$.
Random variables are denoted as capital English letters.
The tilde sign indicates the value of parameters in the last 
iteration of EM, {\it e.g.}, $\tilde{\theta}$.
Given a function $f_{\theta}$, we denote $f_{\tilde{\theta}}$ 
by $\tilde{f}_\theta$ or simply $\tilde{f}$, if the 
parameter $\tilde{\theta}$ is implied. Additional notations, as summarized in Table~\ref{tab:symbol},
will be explained in more details later. 

\begin{table}[ht!] 
\caption{Symbols and descriptions of parameters, random variables, and statistics.}
\begin{tabular}{c | l}
\hline
Symbols & Descriptions \\\hline
$O_i$ & subject $i$ \\
$\tau_i$ & rate of subject reliability\\
$\alpha_i,\beta_i$ & shape of subject regularity\\
$\gamma$ & rate of agreement by chance\\
${\Theta}$ & union of parameters \\
$T_j^{(k)}$ & whether $O_j$ reliably response \\
$J_i^{(k)}$ & rate of $O_i$ agreeing with 
other reliable responses \\
$I_{i,j}^{(k)}$ & whether $O_i$ agrees with the responses from $O_j$\\
$\omega_i^{(k)}(\cdot)$ & cumulative degree of
responses agreed by $O_i$ \\
$\psi_i^{(k)}(\cdot)$ & cumulative degree of responses \\
$r_j^{(k)}(\cdot)$ & a ratio amplifies or discounts the reliability of $O_j$\\
$\tilde{\tau}_i^{(k)}$ & sufficient statistics of posterior $T_i^{(k)}$, given $\tilde{\Theta}$\\
$\tilde{\alpha}_i^{(k)},\tilde{\beta}_i^{(k)}$ & 
 sufficient statistics of posterior $J_i^{(k)}$, given $\tilde{\Theta}$\\
\hline
\end{tabular}
\label{tab:symbol}
\end{table}

\subsection{Agreement Multigraph}
We represent the data as a directed multigraph, which does not assume
a particular type of crowdsourced {response}.
Suppose we have prepared $m$ questions in the study, the answers can be binary, 
categorical, ordinal, and multidimensional.
Given a subject pair $(i,j)$ who are asked to look at the 
$k$-th question, one designs an agreement protocol 
that determines whether the answer from subject $i$ agrees with
that from subject $j$. If subject $i$'s agrees with subject $j$'s on task $k$, 
then we set $I_{i,j}^{(k)}=1$. Otherwise, $I_{i,j}^{(k)}=0$.

In our case, we are given ordinal data from multiple channels, 
we define $I_{i,j}^{(k)}=1$ if (sum of) the percentile difference
between two answers $a_i,a_j\in \{1,\ldots,A\}$ satisfies
\begin{equation}
\dfrac{1}{2}\left\lvert P\left[a_i^{(k)}\right] - P\left[a_j^{(k)}\right] \right\rvert
+\dfrac{1}{2}\left\lvert P\left[a_i^{(k)}\!\!+\!\!1\right] - P\left[a_j^{(k)}\!\!+\!\!1\right] \right\rvert \le \delta,
\end{equation} 
The percentile $P[\cdot]$ is calculated from the 
whole pool of answers for each discrete value, and $\delta = 0.2$.
In the above equation, we measure the percentile difference between $a_i$ and $a_j$ as well as that
between $a_i+1$ and $a_j+1$ in order to reduce the effect of imposing discrete values on the answers that are by nature continuous. 
If the condition does not hold, they disagree and $I_{i,j}^{(k)}=0$.
Here we assume that if two scores for the same image 
are within a 20\% percentile interval,
they are considered to reach an agreement. 
Compared with setting a threshold on their absolute difference, such rule
adapts to the non-uniformity of score distribution.
Two subjects can agree with each other by chance or they 
indeed experience similar emotions in response to the same visual stimulus.

{While the choice of the percentile threshold $\delta$ is inevitably subjective, the selection in our experiments was guided by the desire to trade-off the preservation of the original continuous scale of the scores (favoring small values) and a sufficient level of error tolerance (favoring large values). This threshold controls the sparsity level of the multi-graph, and influences the marginal distribution of estimated parameters. Alternatively, one may assess different values of the threshold and make a selection based on some other criteria of preference (if exist) applied to the final results.}

\subsection{Gated Latent Beta Allocation}
This subsection describes the basic probabilistic graphical model we used to 
jointly model subject reliability, which
is independent from the supplied questions, and regularity. We refrain from carrying out a full
Bayesian inference because it is impractical to end users. Instead, we use the mode(s) of the posterior as point estimates.

\begin{figure}[ht!]
\centering
\includegraphics[width=0.35\textwidth]{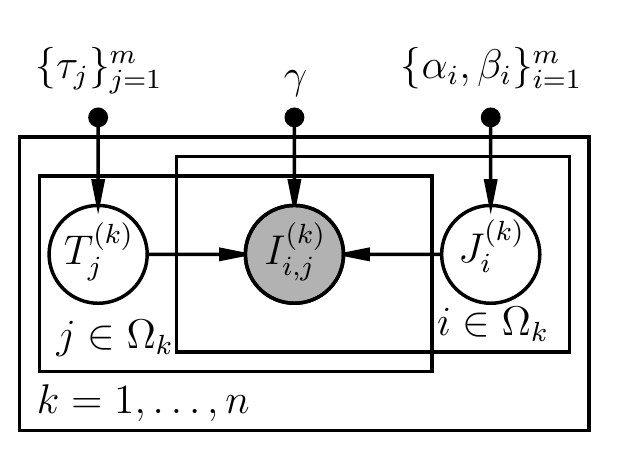}
\caption{Probabilistic graphical model of the proposed Gated Latent Beta Allocation.}
\label{fig:pgm}
\end{figure}

We assume each subject $i$ has a reliability parameter $\tau_{i} \in [ 0,1 ]$
and regularity parameters $\alpha_{i}$, $\beta_{i} > 0$ characterizing his or her agreement
behavior with the population, for $i=1, \ldots ,m$. We also use parameter $\gamma$ for the rate of agreement between subjects out of pure chance. Let $\Theta = ( \{ \tau_{i} , \alpha_{i} ,
\beta_{i} \}_{i=1}^{m} , \gamma )$ be the set of {parameters}. Let $\Omega_{k}$
be the a random sub-sample from subjects $\{ 1, \ldots ,m \}$ who labeled the
stimulus $k$, where $k=1, \ldots ,n$. We also assume sets
$\Omega_{k}$'s are created independently from each other. For each image $k$, every subject
pair from $\Omega_{k}^{2}$, {\it i.e.}, $( i,j )$ with $i \neq j$, has a binary indicator
$I_{i,j}^{( k )} \in \{ 0,1 \}$ coding whether their opinions agree on the
respective stimulus. We assume $I_{i,j}^{( k )}$ are generated from the
following probabilistic process with two latent variables. The first latent variable $T_j^{(k)}$ indicates whether subject $O_j$ is reliable or not. Given that it is binary, a natural choice of model is the Bernoulli distribution.  The second latent variable $J_i^{(k)}$, lying between 0 and 1, measures the extent subject $O_i$ agrees with the other reliable responses. We use Beta distribution parameterized by $\alpha_i$ and $\beta_i$ to model $J_i^{(k)}$ because it is a widely used parametric distribution for quantities on interval $[0,1]$ and the shape of the distribution is relatively flexible. In a nutshell, $T_{j}^{( k )}$ is a latent switch (aka, gate) that controls whether $I_{i,j}^{( k )}$ can be used for the posterior inference
of the latent variable $J_{i}^{( k )}$. Hence, we call our model {\em Gated Latent Beta Allocation} (GLBA). A graphical illustration of the model is shown in Fig.~\ref{fig:pgm}. 

We now present the mathematical formulation of the model. 
For $k=1, \ldots ,n$, we generate 
a set of random variables independently via
\begin{eqnarray}
  T_{j}^{( k )}   \hspace{1em} i.i.d. & \sim & \tmop{Bernoulli} ( \tau_{j} )
  \nocomma , \hspace{1em} j \in \Omega_{k} \;,\\
  J_{i}^{( k )} \hspace{1em} i.i.d. & \sim & \tmop{Beta} ( \alpha_{i} ,
  \beta_{i} ) , \hspace{1em} i \in \Omega_{k} \;,\\ 
  I_{i,j}^{( k )} \left\lvert T_{j}^{(k)}, J_{i}^{(k)} \right. & \sim & \left\{\begin{array}{ll}
    \tmop{Bernoulli} \left( J_{i}^{( k )} \right) \nocomma & \tmop{if}  T_{j}^{( k )}
    =1\\
    \tmop{Bernoulli} ( \gamma ) & \tmop{if}  T_{j}^{( k )} =0
  \end{array}\right. \label{eq:Iijk}
\end{eqnarray}
where the last random process holds for any $j \in \Omega_{k}^{\neg i} \assign
\Omega_{k} - \{ i \}$ and $i \in \Omega_{k}$ with $k=1, \ldots ,n$, and
$\gamma$ is the rate of agreement by chance if one of $i,j$ turns out to be unreliable. 
Here $\{ I_{i,j}^{( k )} \}$ are observed data. 

If a spammer is in the subject pool, his or her reliability parameter $\tau_i$ is zero,
though others can still agree with his or her answers by chance at rate $\gamma$.
On the other hand, if one is very reliable yet often provides controversial
answers, his reliability $\tau_i$ can be one, while he typically disagrees 
with others, indicated by his high irregularity $\mathbb E[J_i^{(k)}] = \frac{\alpha_i}{\alpha_i+\beta_i}\approx 0$. We are interested in finding both types of subjects. However, most of
subjects lie in between these two extremes. 

As an interesting note, Eq.~\eqref{eq:Iijk} is asymmetric, meaning that
$I_{i,j}^{(k)}\neq I_{j,i}^{(k)}$ is possible, a scenario that should never occur by definitions of the two quantities.  We propose to achieve symmetry in the final model by
using the conditional distribution of $I_{i,j}^{(k)}$ and $I_{j,i}^{(k)}$ given that 
$I_{i,j}^{(k)}=I_{j,i}^{(k)}$, and call this model the symmetrized model.
With details omitted, we state that conditioned on $T_i^{(k)}$, $T_j^{(k)}$, $J_i^{(k)}$, and $J_j^{(k)}$, the symmetrized model is still a Bernoulli distribution: 
\begin{equation}
\begin{split}
&I_{i,j}^{( k )} \sim \\
&\tmop{Bernoulli} \left( H\left(\left(J_{i}^{( k )}\right)^{T_i^{(k)}}\gamma^{1-T_i^{(k)}}, 
 \left(J_{j}^{( k )}\right)^{T_j^{(k)}} \gamma^{1-T_j^{(k)}}\right) \right),
\end{split}
\end{equation}
where 
\begin{equation}
H(p,q)=\dfrac{pq}{pq + (1-p)(1-q)}\;.\nonumber
\end{equation}
We tackle the inference and estimation of the asymmetric model for simplicity.

\subsection{Variational EM}
Variational inference is an optimization based strategy for approximating
posterior distribution in complex distributions~\cite{jordan1999introduction}.
Since the full posterior is highly intractable, we consider to use variational
EM to estimate the parameters $\Theta=( \{ \tau_{i} , \alpha_{i} ,
\beta_{i} \}_{i=1}^{m} , \gamma)$~\cite{bernardo2003variational}. 
The parameter $\gamma$ is assumed to be pre-selected by the user 
and does not need to be estimated. To regularize the other parameters in estimation, we use
the empirical Bayes approach to choose priors. Assume the following priors
\begin{eqnarray}
\tau_i &\sim & \mbox{Beta}(\tau_0, 1-\tau_0)\;,\\
\alpha_i + \beta_i & \sim & \mbox{Gamma}(2, s_0) \; .
\end{eqnarray}
By empirical Bayes, $\tau_0$, $s_0$ are adjusted.  
For the ease of notations, we define two auxiliary functions $\omega_{i}^{( k )} ( \cdot )$ and
$\psi_{i}^{( k )} ( \cdot )$:
\begin{equation}
  \omega_{i}^{( k )} ( x ) \assign \sum_{j \in \Omega_{k}^{\neg i}} x_{j}
  I_{i,j}^{( k )} , \hspace{1em} \psi_{i}^{( k )} ( x ) \assign \sum_{j \in
  \Omega_{k}} x_{j}\;.
\end{equation}
Similarly, we define their siblings 
\begin{equation}
\bar{\omega}_{i}^{( k )} ( x ) 
= \omega_{i}^{( k )} ( 1-x ) , \hspace{1em}
\bar{\psi}_{i}^{( k )} ( x ) = \psi_{i}^{( k )} ( 1-x )\;.
\end{equation}
We also define the auxiliary function $r_{j} ( \cdot )$ as
\begin{equation}
  r_{j}^{( k )} ( x ) = \prod_{i \in \Omega_{k}^{\neg j}} \left(
  \dfrac{x_{i}}{\gamma} \right)^{I_{i,j}^{( k )}} \left( \dfrac{1-x_{i}}{1-
  \gamma} \right)^{1-I_{i,j}^{( k )}} .
\end{equation}\\

Now we define the full likelihood function:
\begin{multline}
  L_{k} ( \Theta ;T^{( k )} ,J^{( k )} ,I^{( k )} ) \assign  \prod_{j \in
  \Omega_{k}} \left( ( \tau_{j} )^{T_{j}^{( k )}} ( 1- \tau_{j} )^{1-T_{j}^{( k )}}
  \right)  \\ 
\cdot  \prod_{i \in \Omega_{k}} \dfrac{\left( J_{i}^{( k )}
  \right)^{\alpha_{i}^{( k )}} \left( 1-J_{i}^{( k )} \right)^{\beta_{i}^{( k )}} \phi_{i}^{( k
  )}}{B ( \alpha_{i} , \beta_{i} )} \;,\label{eq:likelihood}
\end{multline}
where auxiliary variables simplifying the equations are
\begin{eqnarray*}
\alpha_{i}^{( k )} &=& \alpha_{i} + \omega_{i}^{( k )} \left( T^{( k )} \right)\;,\\
\beta_{i}^{( k )} &=& \beta_{i} + \psi_{i}^{( k )} - \omega_{i}^{( k )} \left( T^{(
k )} \right)\;,\\
  \phi_{i}^{( k )} &=& \gamma^{\bar{\omega}_{i}^{( k )} \left( T^{( k )} \right)} ( 1-
  \gamma )^{\bar{\psi}_{i}^{( k )}\left( T^{( k )} \right) - \bar{\omega}_{i}^{( k )} \left(
  T^{( k )} \right)} \;,
\end{eqnarray*}
and $B ( \cdot , \cdot )$ is the Beta function. Consequently, 
assume the prior likelihood is $L_{\Theta}(\Theta)$, the 
MAP estimate of $\Theta$ is to minimize 
\begin{equation}
  L ( \Theta ;T,J,I ) \assign L_{\Theta}(\Theta) \prod_{k=1}^{n} L_{k} ( \Theta ;T^{( k )} ,J^{(
  k )} ,I^{( k )} )\;.
\end{equation}
We solve the estimation using variational EM method with a fixed $(\tau_0, s_0)$ and varying $\gamma$.
The idea of variational methods is to approximate the posterior by a factorizable
template, whose probability distribution minimizes its KL divergence to the
true posterior. Once the approximate posterior is solved, it is then used in
the E-step in the EM algorithm as the alternative to the true posterior. The
usual M-step is unchanged. Each time $\Theta$ is estimated,
we adjust prior $(\tau_0, s_0)$ to match the mean of the MAP estimates
of $\{\tau_i\}$ and $\left\{\dfrac{\alpha_i + \beta_i}{2}\right\}$ respective 
until they are sufficiently close.

\tmtextbf{E-step}. We use the factorized Q-approximation with variational
principle: 
\begin{equation}
p_{\Theta} \left( T^{( k )} ,J^{( k )} \left\lvert I^{( k )} \right.\nobracket \right) \approx
  \prod_{j \in \Omega_{k}} q^{\ast}_{T_{j} , \Theta} \left( T_{j}^{( k )} \right)
  \prod_{i \in \Omega_{k}} q^{\ast}_{J_{i} , \Theta} \left( J_{i}^{( k )} \right) .
\end{equation}
\begin{itemizedot}
  \item Let 
  \begin{equation}~\label{eq:qt}
  \begin{split}
 & q^{\ast}_{T_{j},\Theta} \left( T_{j}^{( k )} \right) \propto \\
 & \exp
  \left( \mathbbm{E}_{J,T^{\neg j}} \left[ \log  L_{k} \left( \Theta ;T^{( k )} ,J^{( k )}
  ,I^{( k )} \right) \right] \right)\;, 
\end{split}
\end{equation}
whose distribution can be written as 
  \[\tmop{Bernoulli}
  \left( \dfrac{\tau_{j} R_{j}^{( k )}}{\tau_{j} R_{j}^{( k )} +1-
  \tau_{j}} \right),\] where $\log R_j^{(k)}=\mathbb E_{J}\left[\sum_{i\in \Omega_k^{\neg j}} \log \left(r_i^{(k)} (J^{(k)})\right)\right]$.
  As suggested by Johnson and Kotz~\citep{johnson1995chapter}, the geometric mean can be numerically approximated by
 \begin{equation}
R_{j}^{( k )} \approx \prod\limits_{i\in\Omega_k^{\neg j}} \dfrac{1}{\alpha_i^{(k)} + \beta_i^{(k)}}
  \left(\dfrac{\alpha_i^{(k)}}{\gamma}\right)^{I_{i,j}^{(k)}}\left(\dfrac{\beta_i^{(k)}}{1-\gamma}\right)^{1-I_{i,j}^{(k)}},
  \label{eq:Rapprox}
\end{equation}
if both $\alpha_i^{(k)}$ and $\beta_i^{(k)}$ are sufficiently larger than 1. 
    
  \item Let 
\begin{equation}
  \begin{split}
 & q^{\ast}_{J_{i},\Theta} ( J_{i}^{( k )} ) \propto \\
 & \exp
  \left( \mathbbm{E}_{T,J^{\neg i}} \left[ \log  L_{k} \left( \Theta ;T^{( k )} ,J^{( k )}
  ,I^{( k )} \right) \right] \right)\;,
  \end{split}
\end{equation}
  whose distribution is
  \[ \tmop{Beta} ( \alpha_{i} + \omega_{i}^{( k )} ( \tau ) , \beta_{i} +
     \psi_{i}^{( k )} ( \tau ) - \omega_{i}^{( k )} ( \tau ) )\; . \]
\end{itemizedot}
Given parameter $\tilde{\Omega} = \{ \tilde{\tau}_{i} , \tilde{\alpha}_{i} ,
\tilde{\beta}_{i} \}_{i=1}$, we can compute the approximate posterior
expectation of the log likelihood, which reads 
\begin{eqnarray}
&&\mathbbm{E}_{T,J |
\tilde{\Theta} ,I \nobracket} \log  L_{k} ( \Theta ;T^{( k )} ,J^{( k )} ,I^{(
k )} ) \approx \nonumber \\ 
&&  \tmop{const} \nosymbol .+ \log L_{\Theta} (\Theta) + \nonumber \\ 
&&  \sum_{j \in \Omega_{k}} \left( \tilde{\tau}_{i}^{( k )} \log
  \tau_{j} + ( 1-\tilde{\tau}_{i}^{( k )} ) \log ( 1- \tau_{j} ) \right) + \nonumber \\
&&  \sum_{i \in \Omega_{k}} \left\langle \left(\begin{array}{c}
    \alpha_{i} \\
 \beta_{i}
  \end{array}\right) , \dfrac{\nabla B ( \tilde{\alpha}_{i}^{( k )} ,
  \tilde{\beta}_{i}^{( k )} )}{B ( \tilde{\alpha}_{i}^{( k )} ,
  \tilde{\beta}_{i}^{( k )} )} \right\rangle - \nonumber \\ 
&&  \sum_{i \in \Omega_{k}} \log  B
  ( \alpha_{i} , \beta_{i} ) + 
 \log \gamma \sum_{i \in \Omega_{k}}
  \bar{\omega}_{i}^{( k )} \left( \tilde{\tau}_{i}^{( k )} \right) + \nonumber \\
&&  \log ( 1- \gamma ) \sum_{i \in
  \Omega_{k}} \left( \bar{\psi}_{i}^{( k )} \left( \tilde{\tau}_{i}^{( k )} \right) - \bar{\omega}_{i}^{(
  k )} \left( \tilde{\tau}_{i}^{( k )} \right) \right) ,
\end{eqnarray}
where relevant statistics are defined as
\begin{eqnarray}
\tilde{\alpha}_{i}^{( k )} &=& \tilde{\alpha}_{i} + \omega_{i}^{( k )} (
\tilde{\tau} )\;, \nonumber \\
\tilde{\beta}_{i}^{( k )} &=& \tilde{\beta}_{i} +
\psi_{i}^{( k )} ( \tilde{\tau} ) - \omega_{i}^{( k )} ( \tilde{\tau} )\;, \mbox{ and } \label{eq:statistics} \\ 
\tilde{\tau}_{i}^{( k )} &=& \dfrac{ \tilde{R}_{i}^{( k )} \tilde{\tau}_{i}}{
\tilde{R}_{i}^{( k )}  \tilde{\tau}_{i} +1- \tilde{\tau}_{i}}\;.\nonumber 
\end{eqnarray}
Remark 
$B(\cdot,\cdot)$ is the Beta function, and $\tilde{R}_{i}^{( k )}$ is calculated
from approximation Eq.~\eqref{eq:Rapprox}

\tmtextbf{M-step}. Compute the partial derivatives of $L$ with respect to $\alpha_{i}$
and $\beta_{i}$: let $\Delta_i$ be the set of images that are labeled by subject $i$.
We set $\partial L / \partial \alpha_i =0 $ and $\partial L / \partial \beta_i = 0$ for each $i$, which reads
\begin{eqnarray}
  && \left(\dfrac{\alpha_i + \beta_i}{s_0} - \log (\alpha_i + \beta_i)\right) \cdot
  \left(\begin{array}{c}
     1\\
     1
  \end{array}\right) \nonumber \\  
   && = \sum_{k \in \Delta_{i}} \dfrac{\nabla B (
  \tilde{\alpha}_{i}^{( k )} , \tilde{\beta}_{i}^{( k )} )}{B (
  \tilde{\alpha}_{i}^{( k )} , \tilde{\beta}_{i}^{( k )} )} - \dfrac{\nabla B
  ( \alpha_{i} , \beta_{i} )}{B ( \alpha_{i} , \beta_{i} )} \nonumber\\
  && = \sum_{k \in \Delta_{i}} \left(\begin{array}{c}
    \Psi ( \tilde{\alpha}_{i}^{( k )} ) - \Psi ( \tilde{\alpha}_{i}^{( k )} +
    \tilde{\beta}_{i}^{( k )} )\\
    \Psi ( \tilde{\beta}_{i}^{( k )} ) - \Psi ( \tilde{\alpha}_{i}^{( k )} +
    \tilde{\beta}_{i}^{( k )} ) 
  \end{array}\right)  \nonumber \\ 
   && \;\;\;\; - | \Delta_{i} | \cdot \left(\begin{array}{c}
    \Psi ( \alpha_{i} ) - \Psi ( \alpha_{i} + \beta_{i} ) \\
    \Psi ( \beta_{i} ) - \Psi ( \alpha_{i} + \beta_{i} )
  \end{array}\right)\;, \label{eq:alphabeta}
\end{eqnarray}
where $\Psi ( x ) \in [ \log ( x-1 ) , \log  x ]$ is the Digamma function. 
The above two equations can be practically solved by Newton-Raphson method 
with a projected modification 
(ensuring $\alpha,\beta$ always are greater than zero).

Compute the derivatives of $L$ with respect to $\tau_{i}$ and set $\partial L / \partial \tau_i = 0$, which reads
\begin{eqnarray}
  \tau_{i} & = & \dfrac{1}{| \Delta_{i} |+1} \left(\tau_0 + \sum_{k \in \Delta_{i}}\tilde{\tau}_{i}^{( k
  )}\right)\;. \label{eq:tau} 
\end{eqnarray}
Compute the derivatives of $L$ w.r.t. $\gamma$ and set to zero, which reads
\begin{equation}
  \gamma = \dfrac{\sum_{i \in \Omega_{k}} \bar{\omega}_{i}^{( k )} ( \tilde{\tau}_{i}^{(
  k )} )}{\sum_{i \in \Omega_{k}} \bar{\psi}_{i}^{( k )} ( \tilde{\tau}_{i}^{( k )} )}\;.
  \label{eq:gamma}
\end{equation}
In practice, the update formula for $\gamma$ needs not to be used if $\gamma$ is pre-fixed. 
See Algorithm~\ref{alg:main} for details. 

\subsection{The Algorithm}

We present our final algorithm to estimate all parameters by
knowing the multigraph data $\{ I_{i,j}^{( k )} \}$. Our algorithm is designed based
on Eqs.~\eqref{eq:alphabeta},~\eqref{eq:tau}, and~\eqref{eq:gamma}.
In each EM iteration, there are two loops: one for collecting
relevant statistics for each subgraph, and the other for re-computing
the parameter estimates for each subject. 
Please refer to Algorithm~\ref{alg:main} for details.

\begin{algorithm}[H]
 \caption{Variational EM algorithm of GLBA}\label{alg:main}
\begin{algorithmic}[1]
 \renewcommand{\algorithmicrequire}{\textbf{Input:}}
 \renewcommand{\algorithmicensure}{\textbf{Output:}}
 \REQUIRE A multi-graph $\{I_{i,j}^{k}\in \{0,1\} \}_{i,j\in \Omega_k}$, $0<\gamma<0.5$
 \ENSURE subject parameters $\Theta=(\{(\tau_i, \alpha_i, \beta_i)\}_{i=1}^m,\gamma)$
 \\ \textit{Initialisation} : $\tau_0=0.5, \alpha_i=\beta_i=\tau_i=1.0,i=1,\ldots,m$
 \REPEAT
  \FOR {$k = 1$ to $n$}
  \STATE compute statistics $\tilde{\alpha}_i^{(k)}, \tilde{\beta}_i^{(k)}, \tilde{\tau}_i^{(k)}$ by Eq.~\eqref{eq:statistics};
  \ENDFOR
  \FOR {$i = 1$ to $m$}
  \STATE solve $(\alpha_i,\beta_i)$ from Eq.~\eqref{eq:alphabeta} (Newton-Raphson);
  \STATE compute $\tau_i$ by Eq.~\eqref{eq:tau};
  \ENDFOR
  \STATE (optional) update $\gamma$ from Eq.~\eqref{eq:gamma};
 \UNTIL $\{(\tau_i, \alpha_i, \beta_i)\}_{i=1}^m$ are all converged. 
 \RETURN $\Theta$ 
\end{algorithmic} 
\end{algorithm}
 
\section{Experiments}\label{sec:exp}

\subsection{Data Sets}


We studied a crowdsourced affective data set acquired from 
the Amazon Mechanical Turk (AMT) platform~\cite{xin2016}.
The affective data set is a collection of image stimuli and their affective labels
including valence, arousal, dominance and likeness (degree of appreciation). 
Labels for each image are ordinal: \{1, ... , 9\} for the first
three dimensions, and \{1, ..., 7\} for the likeness dimension. 
{The study setup and collected data statistics have been detailed in~\citep{xin2016}, which
we describe briefly here for the sake of completeness. }

{At the beginning of a session, the AMT study host provides the subject brief training on the concepts of affective dimensions. Here are descriptions used for
valence, arousal, dominance, and likeness. 
\begin{itemize}
\item Valence: degree of feeling happy vs. unhappy
\item Arousal: degree of feeling excited vs. calm
\item Dominance: degree of feeling submissive vs. dominant
\item Likeness: how much you like or dislike the image
\end{itemize}
The questions presented to the subject for each image are given below in exact wording.
\begin{itemize}
\item Slide the solid bubble along each of the bars associated with the 3 scales (Valence, Arousal, and Dominance) in order to indicate how you ACTUALLY FELT WHILE YOU OBSERVED THE IMAGE.
\item How did you like this image? (Like extremely,
 Like very much,
 Like slightly,
 Neither like nor dislike,
 Dislike slightly,
 Dislike very much,
 Dislike extremely)
\end{itemize}
}

{Each AMT subject is asked to finish a set of labeling tasks,
and each task is to provide affective labels on a single image from a prepared set, called the EmoSet. This set contains
around 40,000 images crawled from the Internet using affective keywords. Each task 
is divided into two stages. First, the subject views the image; and second, he/she provides ratings in the emotion dimensions through a Web interface. Subjects usually spend three to ten seconds to view each image, and five to twenty seconds 
to label it. The system records the time durations respectively for the two stages of each task and calculates the average
cost (at a rate of about 1.4 US Dollars per hour). Around 4,000 subjects were recruited in total.}
For the experiments below, 
we retained image stimuli that have received affective labels from at least four subjects.
Under this screening, the AMT data have 47,688 responses from 
2,039 subjects on 11,038 images. Here, one response refers to the labeling of one image by one subject conducted in one task.

Because humans can naturally feel differently from each other in their affective experiences, there was no gold standard criterion to identify spammers.
Such a human emotion data set is difficult to analyze and 
the quality of data is hard to assess. Among several emotion dimensions,
we found that participants were more consistent in the valence dimension. 
As a reminder, valence is the rated degree of positivity of emotion
evoked by looking at an image. We call the variance of the ratings from different subjects on the same image the within-task variance, while the variance of the ratings from all the subjects on all the images the cross-task variance. For valence and likeness, the within-task variance accounts for about 70\% of the cross-task variance,
much smaller than for the other two dimensions. 
Therefore, the remaining experiments were focused on 
evaluating the regularity of image valences in the data.

\subsection{Baselines for Comparison}~\label{sec:baseline}
{We discuss below several baseline methods or models with which we compare our method. \\
\textbf{Dawid and Skene~\citep{dawid1979maximum}.} 
Our method falls into the general category of consensus methods in the literature of statistics and machine learning,
where the spammer filtering decision is made completely based on the labels provided by observers. Those consensus methods
have been developed along the line of Dawid and Skene~\citep{dawid1979maximum}, and they mainly deal with categorical labels by
modeling each observer using a designated confusion matrix. More recent developments of the observer models have been 
discussed in~\citep{sheshadri2013square}, where a benchmark has shown that the Dawid-Skene method is still quite
competitive in unsupervised settings according to a number of real-world data sets for which ground-truth labels are believed to exist albeit unknown. However, this method is not directly applicable to our scenario. To enable comparison with this baseline method, 
we first convert each affective dimension into a categorical label by thresholding. 
We create three categories: high, neural, and low, each covering a continuous range of values on the scale. For example, 
high valence category implies a score greater than a neural score ({\it i.e.}, 5) by more than a threshold ({\it e.g.}, 0.5). Such a thresholding approach has 
been adopted in developing affective categorization systems, {\it e.g.}~\citep{datta2006studying,lu2012shape}.\\
\textbf{Time duration.} In the practice of data collection, the host filtered spammers by a simple criterion---to declare a subject spammer if he
spends substantially less time on every task. The labels provided by the identified spammers were then excluded from the data set for subsequent use, and the host also declined to pay for the task.
However, some subjects who were declined to be paid wrote emails to the host arguing for their cases.
Under this spirit, in our experiments, we form a baseline method that uses the average time duration of each subject to red-flag a spammer.\\
\textbf{Filtering based on gold standard examples.} A widely used spammer detection approach in crowdsourcing
is to create a small set with known ground truth labels and use it to spot anyone who gives incorrect labels. 
However, such a policy was not implemented in our data collection process because as we argued earlier, there is simply no ground truth for the emotion responses to an image in a general sense.  On the other hand, just for the sake of comparison, it seems reasonable to find a subset of images that evoke such extreme emotions that ground truth labels can be accepted. This subset will then serve the role of gold standard examples. 
We used our method to retrieve a subset of images which evoke extreme emotions
with high confidence (see Section~\ref{sec:image_conf} for confidence score and emotion score calculation). For the valence dimension,
we were able to identify at most 101 images with valence score $\ge 8$ (on the scale of $1\ldots 9$) with over $90\%$ confidence and 37
images with valence score $\le 2$ with over $90\%$ confidence. We also looked at those images one by one
(as provided in the supplementary materials) and believe that within a reasonable tolerance of doubt those images should evoke clear emotions in the valence dimension. 
Unfortunately, only a small fraction of subjects in our pool have labeled at least one image from this "gold standard" subset.
Among this small group, their disparity from the gold standard enables us to find three susceptible spammers. To see whether these three susceptible spammers can also be detected by our method, we find that their reliability scores $\tau\in[0,1]$ are $0.11, 0.22, 0.35$ respectively. In Fig.~\ref{fig:simulated}, we plot the distribution of $\tau$ of the entire subject pool. These three scores are clearly on the low end with respect to the scores of the other subjects. Thus the three spammers are also assessed to be highly susceptible by our model.}

{In summary, while we were able to compare our method with the first two baselines quantitatively, with results to be presented shortly, 
comparison with the third baseline is limited
due to the way the AMT data were collected~\cite{xin2016}.}

\subsection{Model Setup}
Since our hypotheses included a random agreement ratio $\gamma$ that is pre-selected, we adjusted the parameter $\gamma$ from 0.3 to 0.48 to
see empirically how it affects the result in practice. 

\begin{figure}[ht!]
\centering
\includegraphics[width=0.45\textwidth]{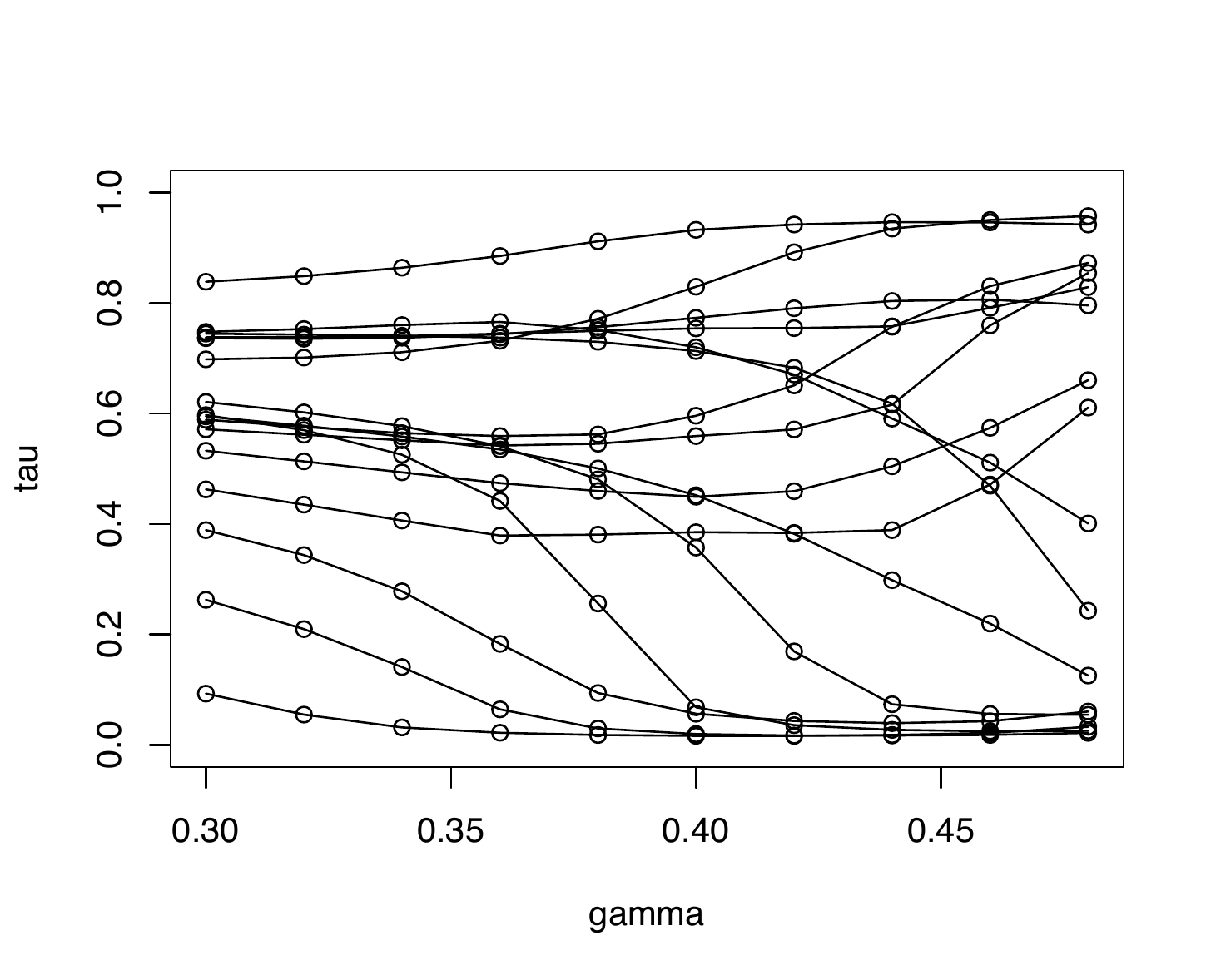}\\[-.4cm]
(a)\\
\includegraphics[width=0.45\textwidth, trim = 0 0 0 1.5cm, clip]{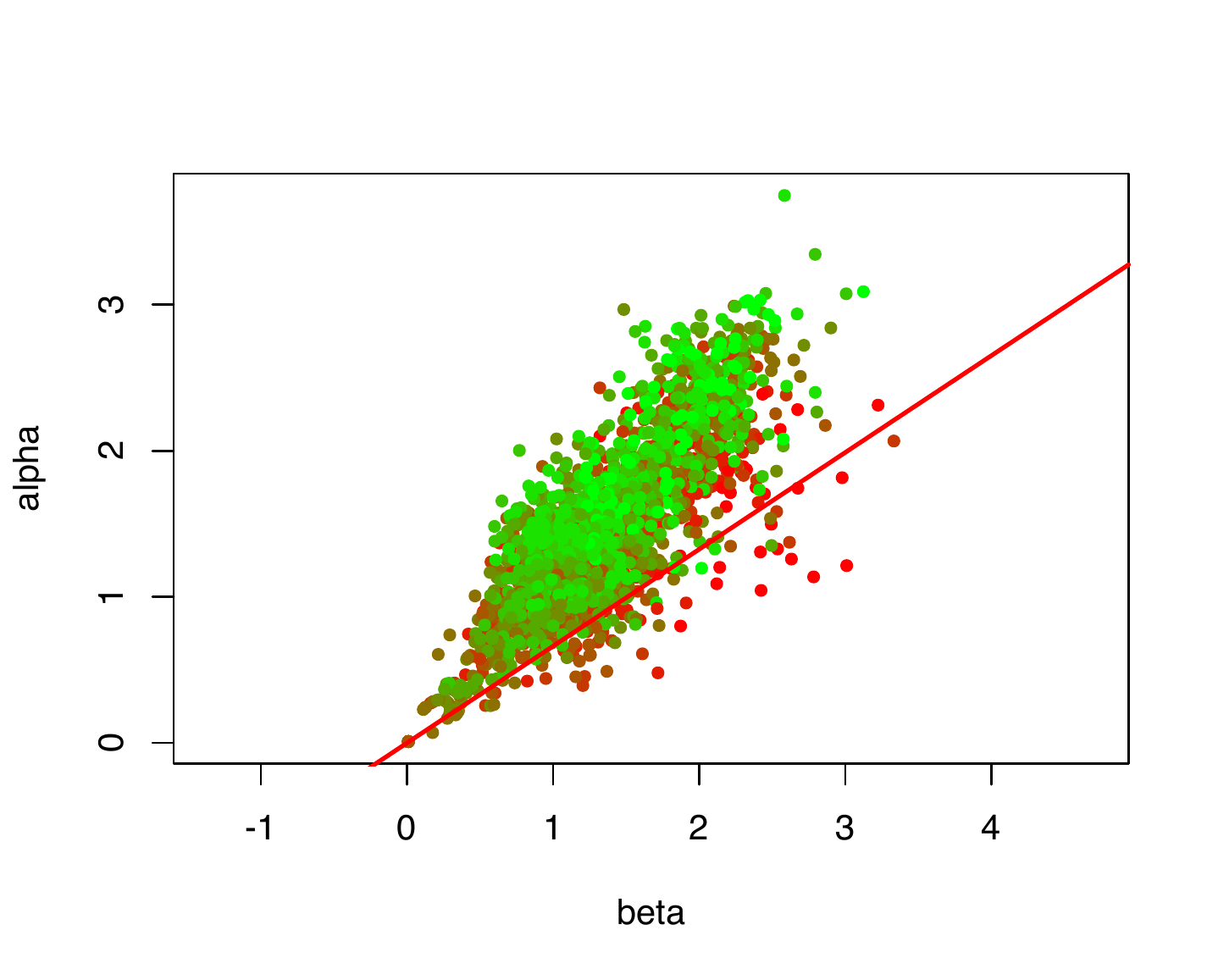}\\[-.4cm]
(b)
\caption{(a) Reliability scores versus $\gamma\in [0.3, 0.48]$ for the top 15 users who provided the most numbers of ratings. 
(b) Visualization of 
the estimated regularity parameters of each worker at a given $\gamma$. Green dots are for workers with high reliability and 
red dots for low reliability. The slope of the red line equals $\gamma$. }
\label{fig:labelers}
\end{figure}

Fig.~\ref{fig:labelers} depicts how the reliability parameter $\tau$ varies with $\gamma$ for different workers in our data set. 
Results are shown for the top 15 users who provided the most numbers of ratings. 
Generally speaking, a higher $\gamma$ corresponds to a higher chance of agreement between workers purely out of random.
From the figure, we can see that a worker providing more ratings is not necessarily more reliable. 
It is quite possible that some workers took advantage of the AMT study
to {earn monetary compensation without paying enough attention to the actual questions}. 

\begin{figure*}[ht!]
\includegraphics[width=0.33\textwidth, trim=0 1.5cm 1cm 0]{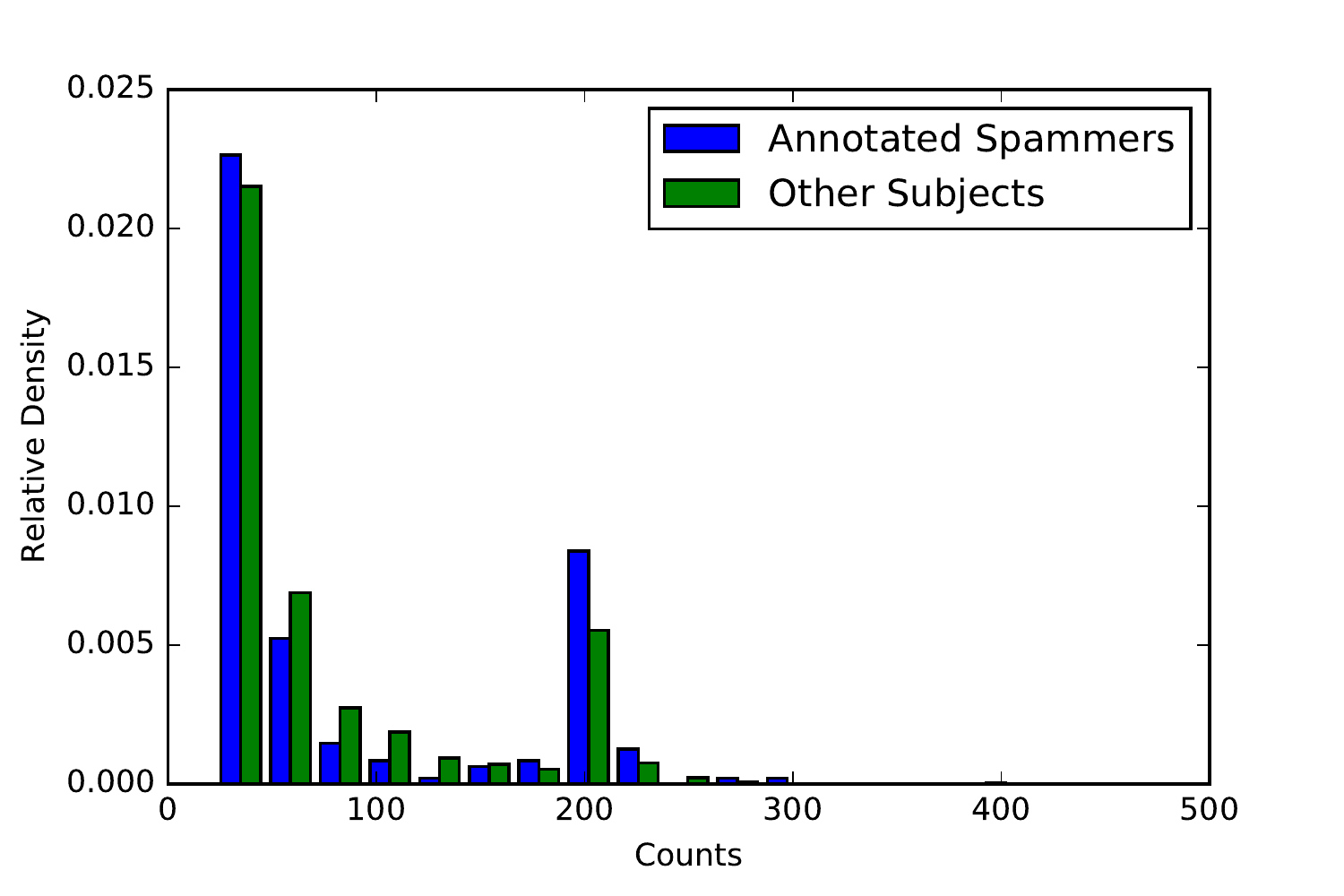}
\includegraphics[width=0.33\textwidth, trim=0 1.5cm 1cm 0]{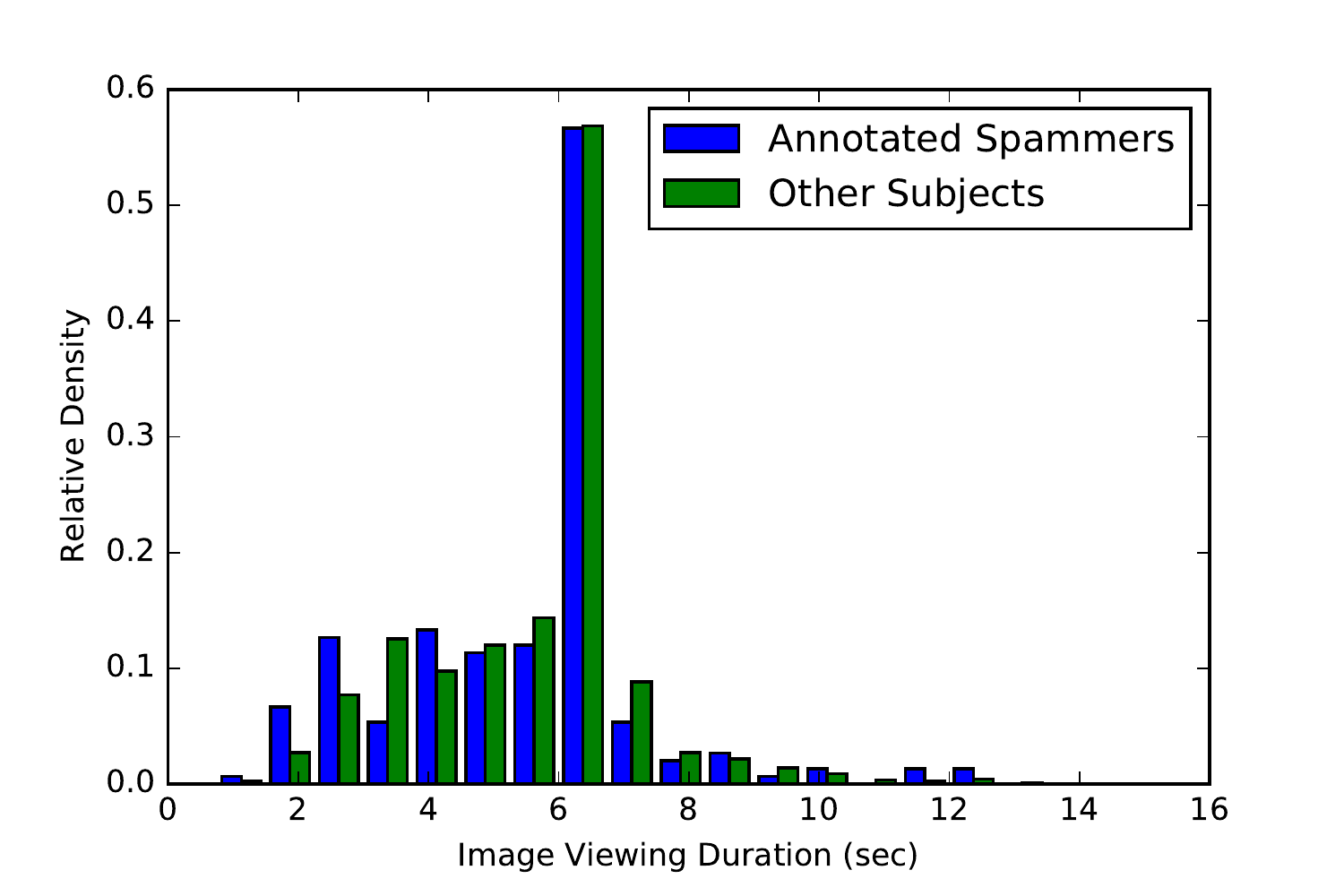}
\includegraphics[width=0.33\textwidth, trim=0 1.5cm 1cm 0]{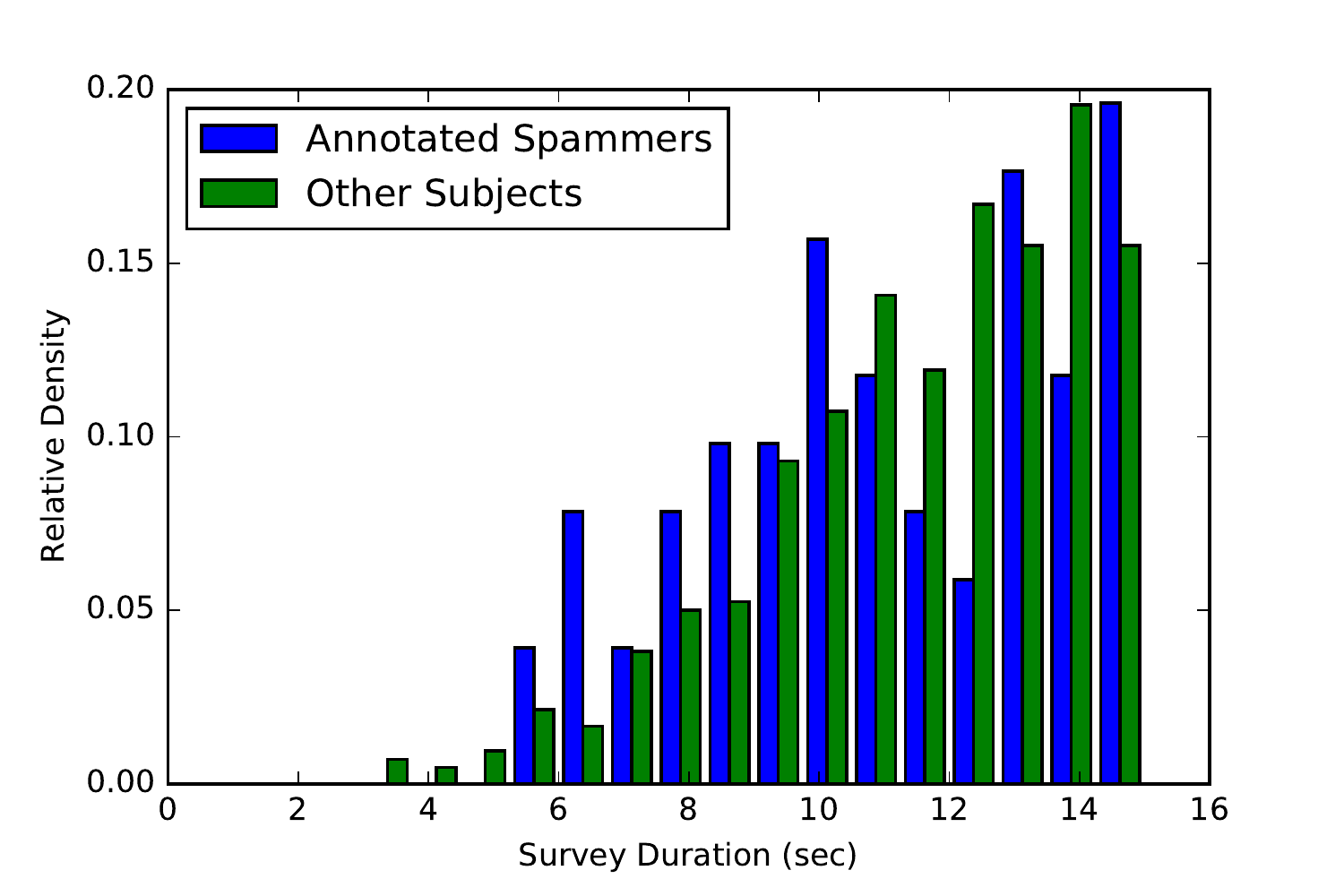}
\caption{Normalized histogram of basic statistics including total number of tasks completed and 
average time duration spent at each of the two stages per task.}\label{fig:hist}
\end{figure*}

\begin{table*}[ht!]
\centering
\colorbox{lightgray}{%
\begin{tabular}{ccc|l}
$\tau_i$ & $\alpha_i$ & $\beta_i$ & reported emotions (sorted)\\\hline\hline
0.19 &1.17&2.43    & \includegraphics[scale=0.5]{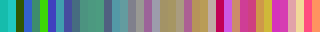}\\
0.08 & 0.75 & 2.20 & \includegraphics[scale=0.5]{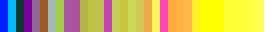}\\
0.08 & 1.16 & 2.50 & \includegraphics[scale=0.5]{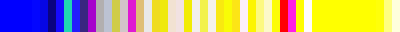}\\
0.09 & 0.67 & 1.70 & \includegraphics[scale=0.5]{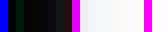}\\
0.03 & 0.94 & 1.90 & \includegraphics[scale=0.5]{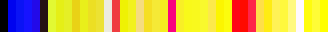}\\
0.17 & 0.72 & 1.47 & \includegraphics[scale=0.5]{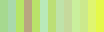}\\
0.06 & 1.14 & 2.50 & \includegraphics[scale=0.5]{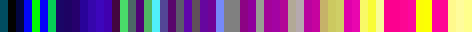}\\
0.17 & 0.86 & 1.79 & \includegraphics[scale=0.5]{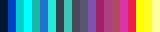}\\
0.04 & 1.01 & 2.63 & \includegraphics[scale=0.5]{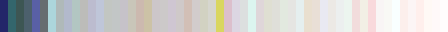}\\
0.03 & 1.08 & 2.84 & \includegraphics[scale=0.5]{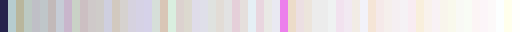}\\\hline\hline
0.92 & 2.29 & 1.49 & \includegraphics[scale=0.5]{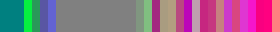}\\
0.94 & 2.55 & 1.98 & \includegraphics[scale=0.5]{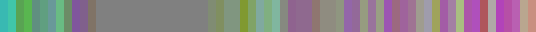}\\
0.95 & 2.61 & 1.68 & \includegraphics[scale=0.5]{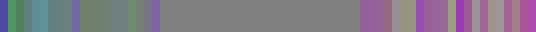}\\
0.92 & 2.40 & 1.66 & \includegraphics[scale=0.5]{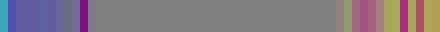}\\
0.91 & 2.21 & 1.40 & \includegraphics[scale=0.5]{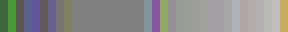}\\
0.92 & 2.45 & 1.97 & \includegraphics[scale=0.5]{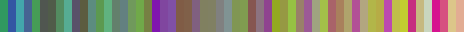}\\
0.93 & 2.38 & 1.69 & \includegraphics[scale=0.5]{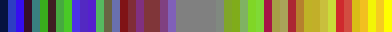}\\
0.93 & 1.76 & 1.40 & \includegraphics[scale=0.5]{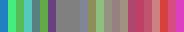}\\
0.91 & 2.44 & 1.86 & \includegraphics[scale=0.5]{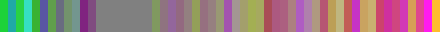}\\
0.92 & 2.30 & 1.85 & \includegraphics[scale=0.5]{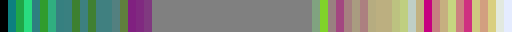}\\
0.92 & 2.45 & 1.82 & \includegraphics[scale=0.5]{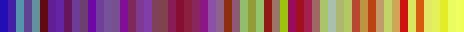}\\
0.91 & 1.64 & 1.29 & \includegraphics[scale=0.5]{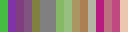}\\
0.90 & 1.68 & 1.12 & \includegraphics[scale=0.5]{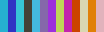}\\
0.91 & 2.72 & 2.22 & \includegraphics[scale=0.5]{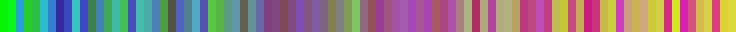}
\end{tabular}
}\caption{Oracles in the AMT data set. 
Upper: malicious oracles whose ${\alpha_i}/{\beta_i}$ is among
the lowest 30, meanwhile $|\Delta_i|$ is greater than 10. 
Lower: reliable oracles whose $\tau_i$ is among the top 30, meanwhile ${\alpha_i}/{\beta_i}>1.2$.
Their reported emotions are visualized by RGB colors. 
The estimates of $\Theta$ is based on the valence dimension.}
\label{fig:visual}
\end{table*}

In Table~\ref{fig:visual}, we demonstrate the valence, arousal, and dominance labels for two categories of subjects.
On the top, the first category contains susceptible spammers with low estimated reliability parameter $\tau$; and on the bottom, the second category contains highly reliable subjects with high values of $\tau$. Each subject takes one row. For the convenience of visualization, we represent the three-dimensional emotion scores given to any image by a particular color whose RGB values are mapped from the values in the three dimensions respectively.  The emotion labels for every image by one subject are then condensed into one color bar. The labels provided by each subject for all his images are then shown as a palette in one row. For clarity, the color bars are sorted in lexicographic order of their RGB values.
One can clearly see that those labels given by the subjects from these two categories exhibit quite different patterns. The palettes of the susceptible spammers are more extreme in terms of saturation or brightness. 
The abnormality of label distributions of the first category naturally
originates from the fact that spammers intended to label the data
by {exerting the minimal efforts and without paying attention} to the questions.

\subsection{Basic Statistics of Manually Annotated Spammers}~\label{sec:spammer}
{For each subject in the pool, by observing all his or her labels in different
emotion dimensions, there was a reasonable chance of spotting abnormality 
solely by visualizing the distribution. If one were a spammer, it often happened that his or her
labels {were highly correlated, skewed or 
deviated in an extreme manner from a neural emotion along different dimensions}. In such cases, it was possible to manually
exclude his or her responses from the data due to his or her high susceptibility.
We applied this same practice to identifying highly susceptible subjects from
the pool. We found about 200 susceptible participants. }

{We studied several basic statistics of this subset in comparison with the whole population:
total number of tasks completed, average time duration spent on image viewing and survey per task. The histograms of these quantities are 
plotted in Fig.~\ref{fig:hist}. One can see that the annotated spammers did not necessarily spend less time or 
finish fewer tasks than the others, and the time duration has shown only marginal sensitivity to those annotated spammers (See Fig.~\ref{fig:hist}). The figures demonstrate that those statistics are not effective criteria for spammer filtering.}

{
We will use this subset of susceptible subjects as a "pseudo-gold standard" set for quantitative comparisons of our method and the baselines in the subsequent studies. As explained previously in \ref{sec:baseline}, other choices of constructing a gold standard set either conflict the high variation nature of emotion responses or yield only a tiny (of size three) set of spammers.   
}

\subsection{Top-K Precision Performance in Retrieving the Real Spammers}
\begin{figure}[htp]
\includegraphics[width=.5\textwidth, trim= 0cm .7cm 0.1cm 2cm, clip]{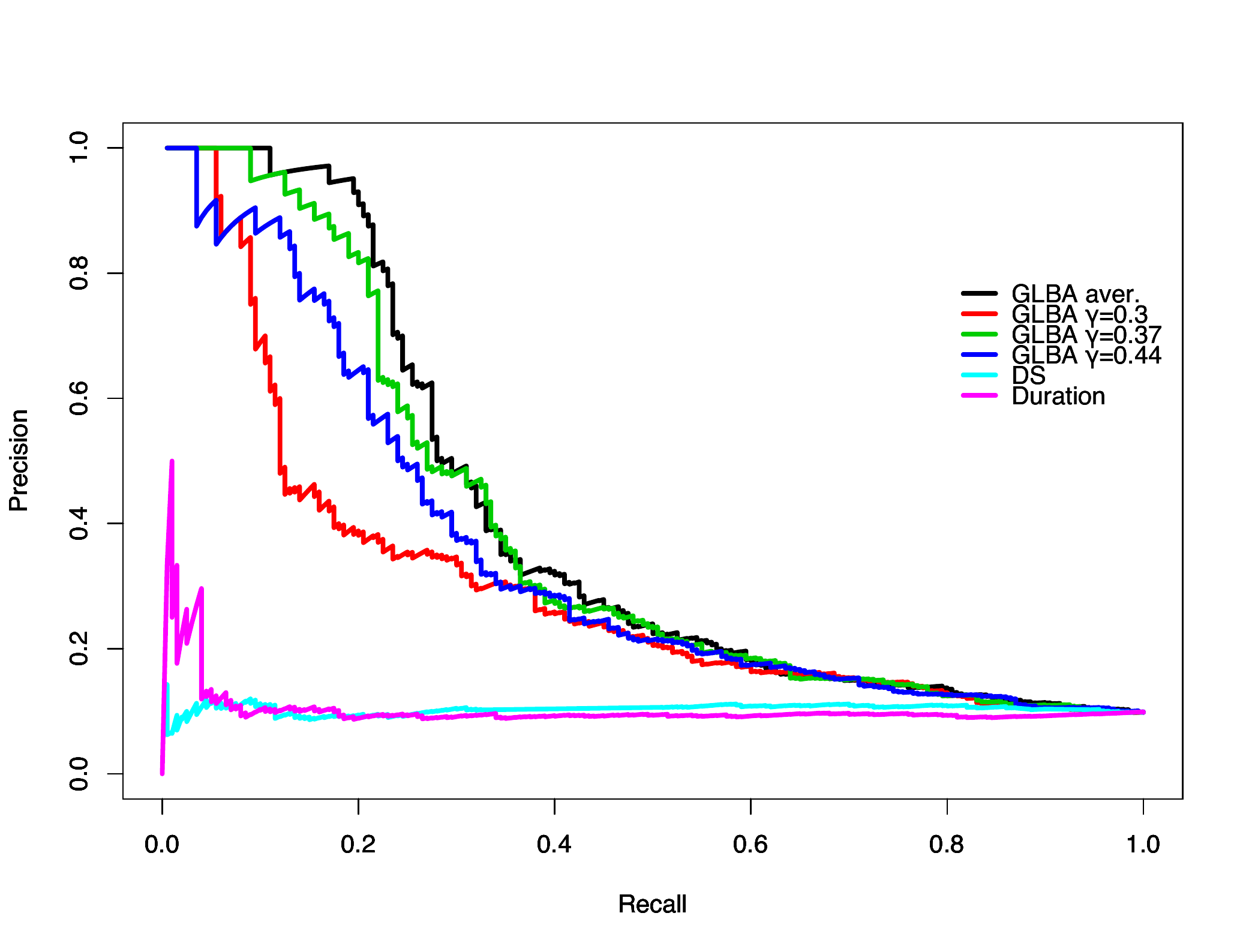}
\caption{The agnostic Precision-Recall curve (by valence) 
based on manually annotated spammers. 
The top 20, top 40 and top 60 precision is 
$100\%$, $95\%$, $78\%$ respectively (black line). 
It is expected that precision drops quickly with increasing recalls,
because the manually annotation process can only identify 
a special type of spammers, while other types of spammers can be identified
by the algorithm. The PR curves at $\gamma=0.3,0.37,0.44$ are also plotted.
{Two baselines are compared: the Dawid and Skene (DS) approach
and the time duration based approach.}}
\label{fig:prc}
\end{figure}
\begin{figure}[htp]
\includegraphics[width=.5\textwidth, trim= 0cm 0.7cm 0.1cm 2cm, clip]{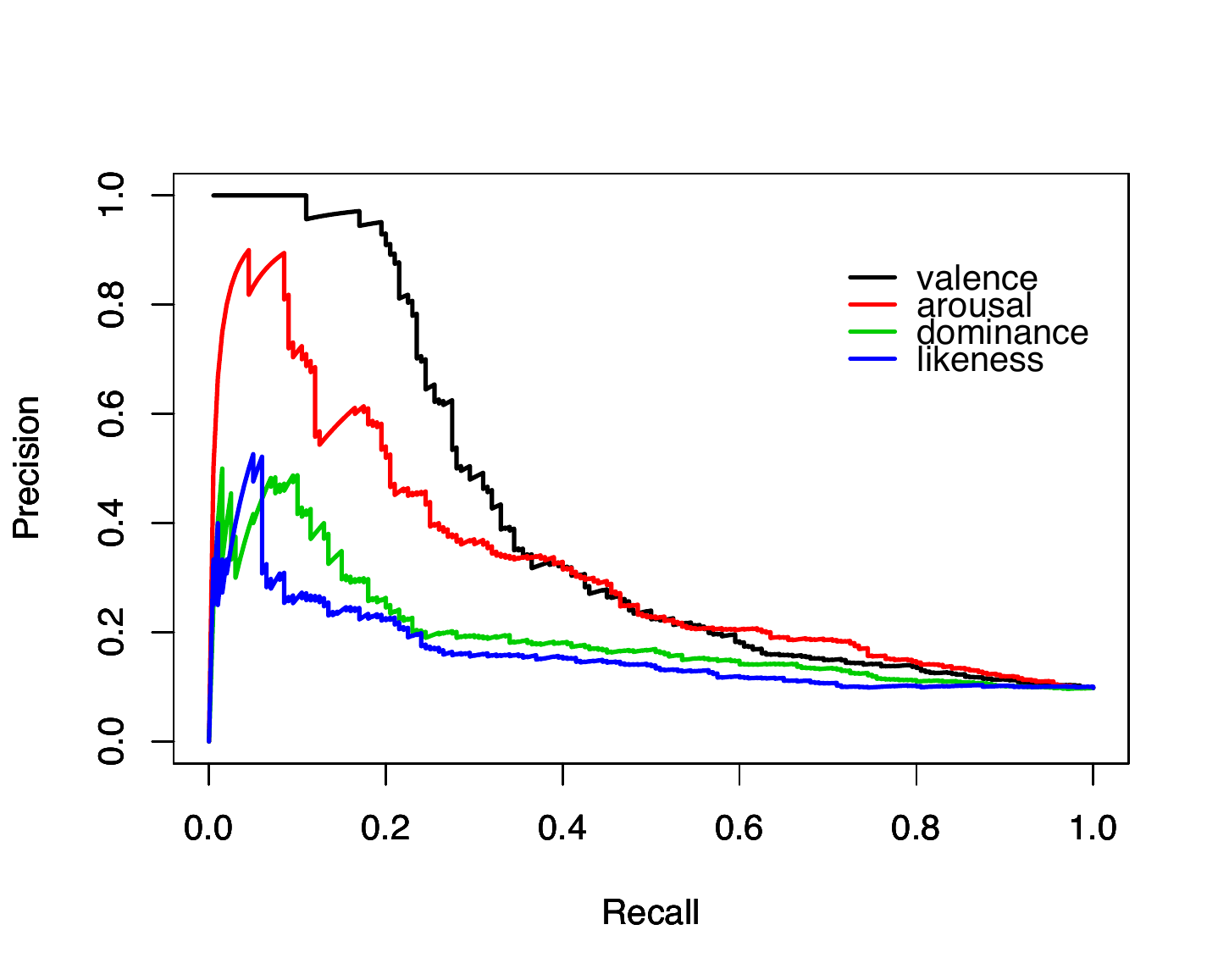}
\caption{The agnostic Precision-Recall curve based on manually annotated spammers
computed from different affective dimensions: valence, arousal, dominance, and likeness.}
\label{fig:prc2}
\end{figure}
We conducted experiments on each affective dimension, and evaluated
whether the subjects with the lowest estimated $\tau$ were supposed to be real spammers {according to the "pseudo-gold standard" subset constructed in Section~\ref{sec:spammer}. }
Since there was no gold standard to correctly classify whether one subject
was truly a spammer or not, we have been agnostic here. 
Based on that subset, we were able to partially evaluate the top-K precision in retrieving the real spammers, especially the most susceptible ones. 

Specifically, we computed the reliability parameter $\tau$ for each subject and chose the $K$ subjects with the lowest values as the most susceptible spammers. Because $\tau$ depends on the random agreement rate $\gamma$, we computed $\tau$'s using 10 values of $\gamma$ evenly spaced out over interval $[0.3, 0.48]$. The average value of $\tau$ was then used for ranking. The Precision Recall Curves
are shown in Fig.~\ref{fig:prc}.
Our method achieves high top-K precision by retrieving 
the most susceptible subjects from the pool according to the average $\tau$. In particular, the top-20 precision
is $100\%$, the top-40 precision is $95\%$, and the top-60 precision
is $78\%$. Clearly, our algorithm
has yielded results well aligned with the human judgment on the most susceptible ones. 
In Fig.~\ref{fig:prc}, we also plot Precision Recall Curves
by fixing $\gamma$ to $0.3,0.37,0.44$ and using the corresponding $\tau$. The result at $\gamma=0.37$
is better than the other two across recalls, indicating that a proper level of
the random agreement rate can be important for achieving the best performance.
{The two baseline methods are clearly not competitive in this evaluation.
The Dawin-Skene method~\cite{dawid1979maximum}, widely used in processing crowdsourced data with objective ground truth labels,
drops quickly to a remarkably low precision even at a low recall. The time duration method, used in the practice of AMT host,
is better than the Dawin-Skene method, yet substantially worse than the performance of our method.}

We also tested this same method of identifying spammers using affective dimensions other than valence. As shown in Fig.~\ref{fig:prc2}, the two most discerning dimensions were valence and arousal.
{It is not surprising that people can reach relatively higher consensus 
when rating images by these two dimensions than by dominance or likeness.
Dominance is much more likely to draw on evidence from context and social situation in most circumstances 
and hence less likely to have its nature determined to a larger extent by the stimulus itself.}

\subsection{Recall Performance in Retrieving the Simulated Spammers}
The evaluation of top-K precision was limited in two respects: (1) the susceptible 
subjects were identified because we could clearly observe their abnormality in terms of
the multivariate distribution of provided labels. 
If the participant labeled the data by acting exactly 
the same as the distribution of the population, we could not manually identify him/her 
using the aforementioned methodology. (2) We still need to determine
if one is a spammer, how likely we are to spot him/her. 

\begin{figure}[htp]
\includegraphics[width=0.5\textwidth]{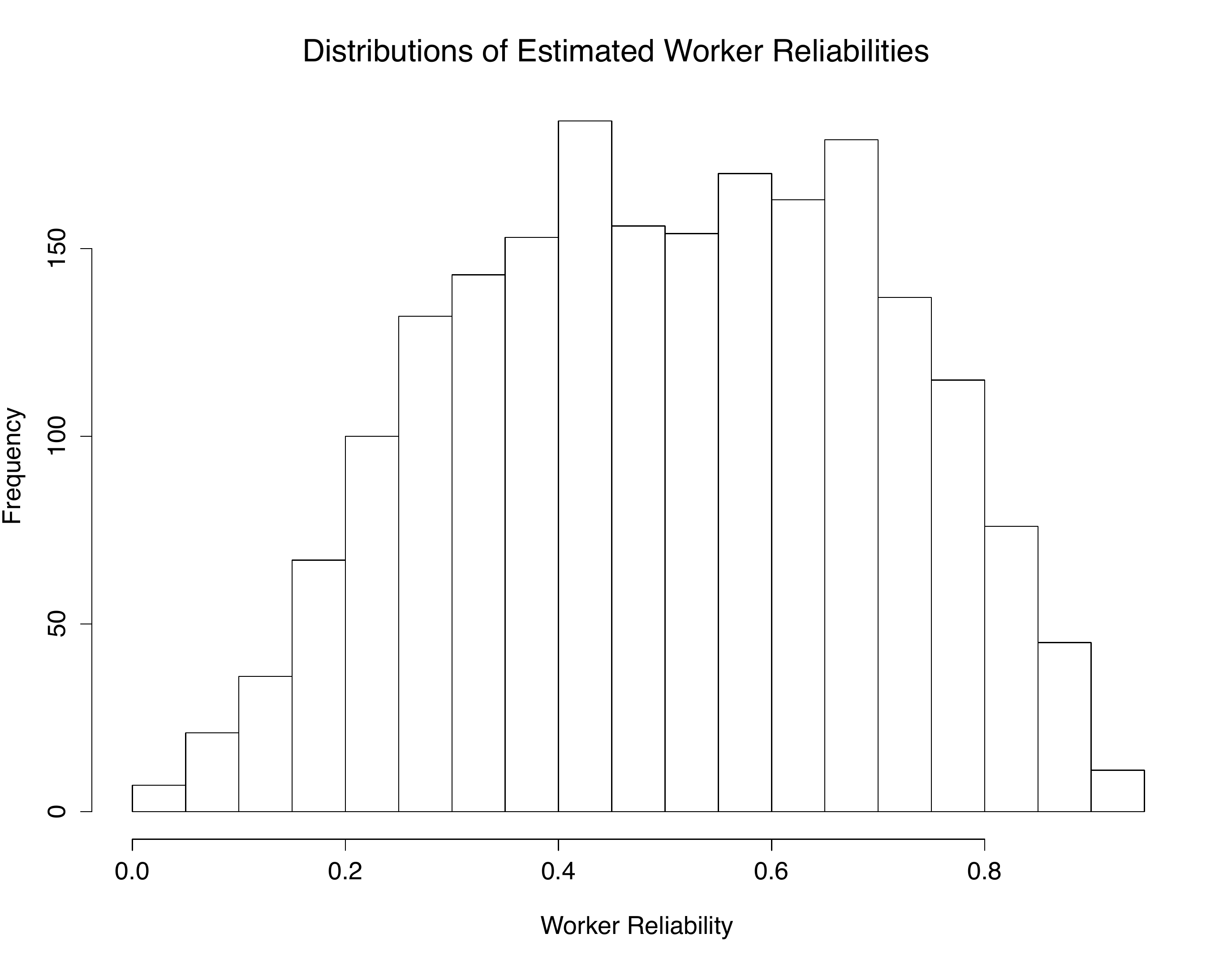}
\begin{tabular}{|c|c|c|c|c|c|}\hline 
Ranges & $\le$ 0.2 & 0.2$\sim$0.4 & 0.4$\sim$0.6 & 0.6$\sim$0.8 & $\ge$0.8\\\hline
Counts & 54 & 34 & 12 & 0 & 0 \\\hline
\end{tabular}
\caption{The histogram distribution of estimated worker reliabilities $\tau$
and statistics of simulated spammers based on 10 repeated runs, each with 10 spammers injected.}
\label{fig:simulated}
\end{figure}

In this study, we simulated several highly ``intelligent'' spammers, who labeled the data
by exactly following the label distribution of the whole population. 
Every time, we generated 10 spammers, who randomly labeled 50 images. 
The labels of simulated spammers were not overlapping. 
We mixed those labels of the simulated spammers 
with the existing data set, and then conducted our method again to determine 
how accurate our approach was with respect to finding the simulated spammers. 
We repeated this process 10 times in order to estimate the $\tau$ 
distribution of the simulated spammers. Results are reported Fig.~\ref{fig:simulated}.
We drew the histogram of the estimated reliability of all real workers
and compared them to the estimated reliability of simulated spammers (in the table included in Fig.~\ref{fig:simulated}).
We noted that more than half of the simulated spammers were identified as highly susceptible
based on the $\tau$ estimation ($\le 0.2$), and none of them were supposed 
to have a high reliability score ($\ge 0.6$). This result validates that our method
is robust enough to spot the ``intelligent'' spammers, even if they disguise themselves
as random labelers within a population.

\subsection{Qualitative Comparison Based on Controversial Examples}\label{sec:image_conf}
\begin{figure}
\includegraphics[width=0.5\textwidth, trim=1cm .5cm 1cm 2cm, clip]{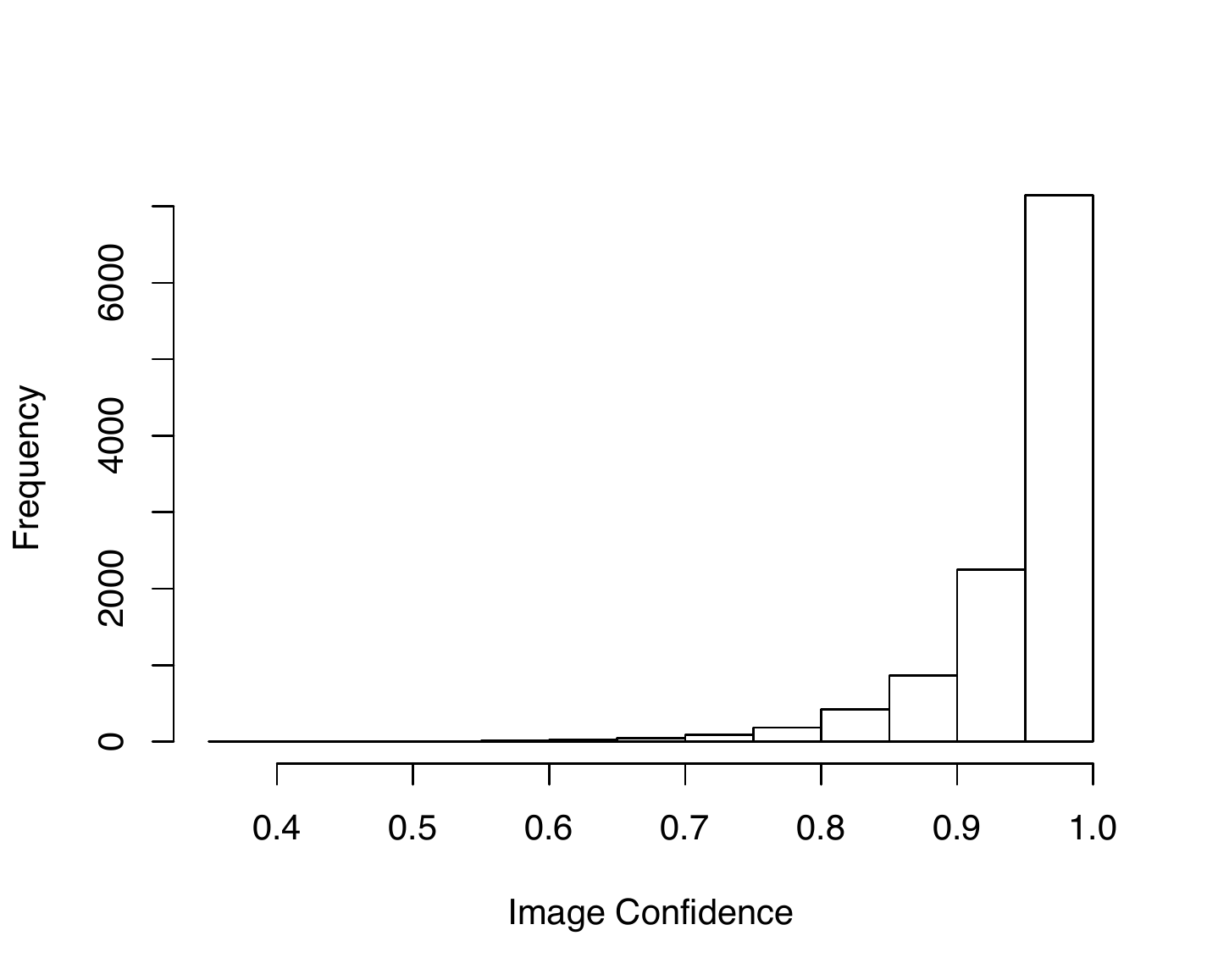}
\caption{The histogram of image confidences estimated based on our method.
About 85\% of images have a confidence scores higher than 90\%.}\label{fig:imageconf}
\end{figure}
To re-rank the emotion dimensions and likenesses of stimuli with the reliability of the subject accounted for, we adopted the following formula to 
find the stimuli with ``reliably'' highest ratings.  Assume each rating $a_i\in [0,1]$. We define the following to replace the usual average:
\begin{equation}
b_k \assign \underbrace {\dfrac{\sum_{i\in\Omega_k} \tau_i a_i^{(k)}}{\sum_{i\in\Omega_k} \tau_i} }_\text{est. score}
\cdot \underbrace{\left(1-\prod_{i\in\Omega_k} (1-\tau_i)\right)}_\text{confidence},
\end{equation}
where $\left(1-\prod_{i\in\Omega_k} (1-\tau_i)\right)\in [0,1]$ is the {\em cumulative 
confidence score} for image $k$. This adjusted rating $b_k$ not only allows more reliable subjects to play a bigger role via the weighted average (the first term of the product) but also modulates the weighted average by the cumulative confidence score for the image. 
Similarly, in order to find those with ``reliably'' lowest ratings, we replace $a_i^{(k)}$ with
$(1 - a_i^{(k)})$ in the above formula and then still seek for the images with the highest $b_k$'s. 

If $b_k$ is higher than a neutral level, then the emotional response to the image is considered high. Fig.~\ref{fig:imageconf} shows
the histogram of image confidence scores estimated by our method.
More than 85\% of images had acquired a sufficient number of quality labels. 
To obtain a qualitative sense of the usefulness of the reliability parameter $\tau$, we compared our approach with the simple average-and-rank scheme by examining
controversial image examples according to each emotion dimension.  Here, being controversial means the assessment of the average emotion response for an image differs significantly between the methods.
Despite the variability of human nature, the majority of the population
were quite likely to reach consensus for a portion of the stimuli. Therefore,
this investigation is meaningful. In Fig.~\ref{fig:example} and Fig.~\ref{fig:example2}, we show example image stimuli that were recognized to clearly deviate from neutral emotions by
one method but not agreed upon by the other. We skipped stimuli images that were fear inducing, visually annoying or improper. Interested readers can see the complete results in the supplementary material.

\subsection{Cost/Overhead Analysis} 

\begin{figure}[htp]
\includegraphics[width=.5\textwidth, trim = 0cm 2cm 0cm 1cm]{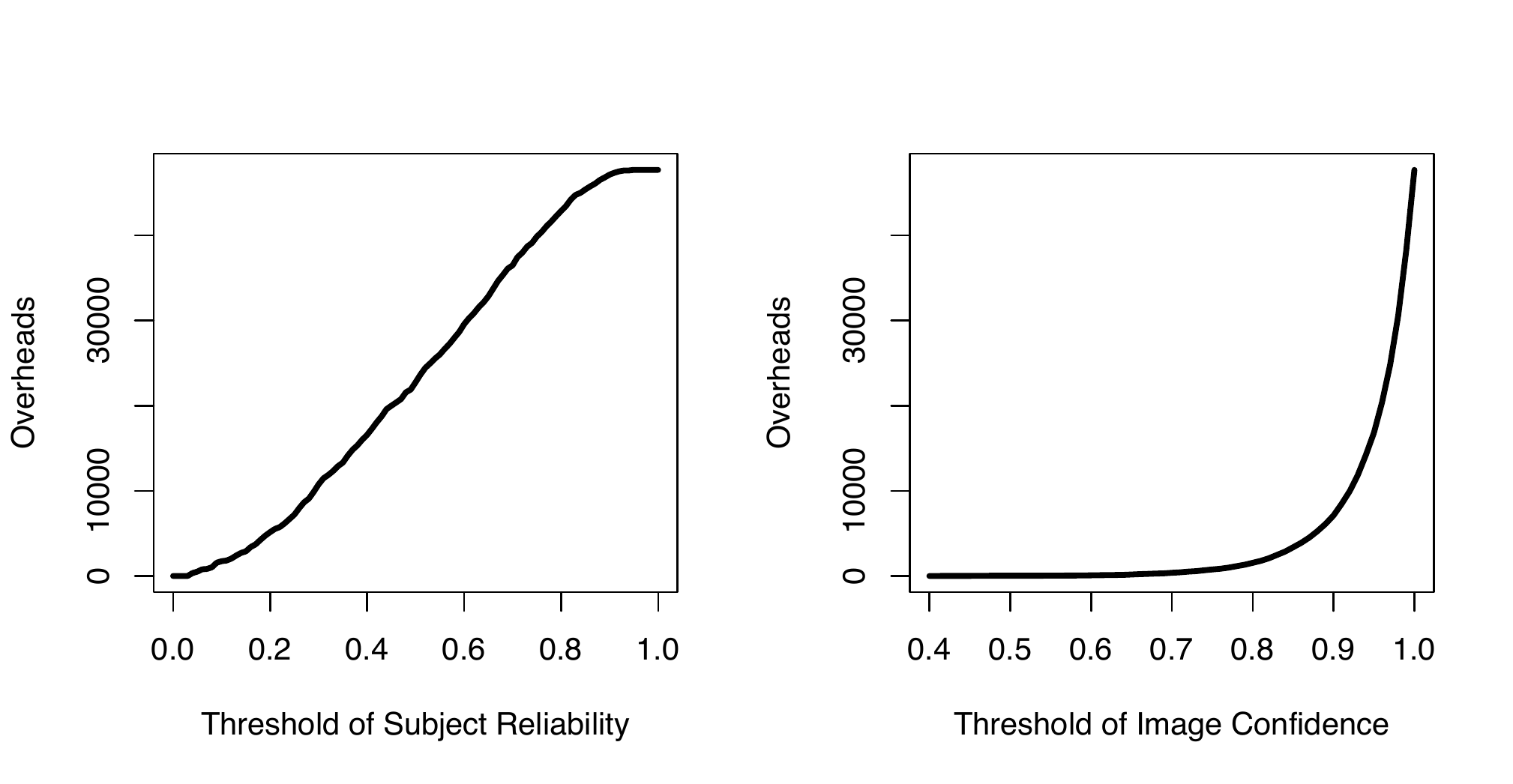}
\caption{Left: Overhead curve based on subject filtering; Right: overhead curve based on image filtering. The overhead is quantified by the number of labels discarded after filtering.}\label{fig:overhead}
\end{figure}

There is an inevitable trade-off between the quality of the labels and the average cost of acquiring them when screening is applied based on reliability. If we set a higher standard for reliability, the quality of the labels retained tends to improve but we are left with fewer labels to use. It is interesting to visualize the trade-off quantitatively.  Let us define overhead numerically as the number of labels removed from the data set when quality control is imposed; and let the threshold on either subject reliability or image confidence used to filter labels be the index for label quality. We obtained what we call {\em overhead curve} in
Figure~\ref{fig:overhead}. On the left plot, the result is based on filtering subjects with reliability scores below a
threshold (all labels given by such subjects are excluded); on the right, it is based on filtering images with confidence scores below a threshold. As shown by the plots,
if either the labels from subjects with reliability scores below 0.3 are discarded or those for images with confidence scores below 90\% are discarded, roughly 10,000 out of 47,688
labels are deemed unusable. At an even higher standard, e.g., subject reliability $\geq .5$ or image confidence level $\geq 95\%$, around half of the labels will be excluded from the data set. Although this means the average per label cost is doubled at the stringent quality standard, we believe the screening is worthwhile in comparison with analysis misled by wrong data. In a large-scale crowdsource environment, it is simply impractical to expect all the subjects to be fully serious. This contrasts starkly with a well-controlled lab environment for data collection. In a sense, post-collection analysis of data to ensure quality is unavoidable. It is indeed a matter of which analysis should be applied. 

\section{Discussions}
{\textbf{Underlying Principles:} 
Our approach to assess the reliability of crowdsourced affective data deviates fundamentally from the standard approaches much concerned with hunting for "ground truth" emotion stimulated by an image.  An individual's emotion response is expected to be naturally different because it depends on subjective opinions rooted in the individual's lifetime exposure to images and concepts, a topic having been pursued long in the literature of social psychology. 
The new principle we adopted here focuses on the relational knowledge about the ratings of the subjects. Our analysis steps away from the use of "ground truth" by recasting the data as relational quantities.}

{
As pointed out by a reviewer, such a relational perspective may be intrinsic in human cognition, going beyond our specific problem here. For instance,
the same spirit of exploiting relationships has already appeared in studies to understand linguistic learning. Gentner~\cite{gentner2010bootstrapping,gentner2010mutual} proposed that one should understand linguistic learning in a relational way. Instead of assuming there are well-formed abstract language concepts to grasp, the human's cognitive ability often starts from analogical processing based on examples of a concept, and then utilizes the symbolic systems (languages) to reinforce and guide the learning, and to facilitate memory of the acquired concepts. The relationships among the examples and the abstract concept play a role in learning hand in hand, refining recursively the understanding of each other. The whole process is an interlocked and repeated improvement of one side assisted by the other. In a similar fashion, our system improves its assessment about which images evoke highly consensus emotion responses and which subjects are reliable. At the beginning, the lack of either kind of information obscures the truth about the other. Or equivalently, knowing either makes the understanding of the other easy. This is a chicken-and-egg situation. Like the proposed way of learning languages, our system pulls out of the dilemma by recursively enhancing the understanding of one side conditioned on what has been known about the other. 
}\\

\noindent{\textbf{Results:} 
We found that the crowdsourced affective data we examined are particularly challenging for the conventional school of observer models, developed along the line of Dawid and Skene~\cite{dawid1979maximum}. We identified two major reasons. First, each image in our data set has a much smaller number of observers, compared with what are typically studied in the benchmarks~\cite{sheshadri2013square}. In our data set, most images were only labeled by 4 to 8 subjects, while many existing benchmark data sets
have tens of subjects per task. Second, a more profound reason is that most images do not have a ground truth affective label at the first place.
This can render ineffective many statistical methods which model the user-task confusion matrix and hence count on the existence of "true" labels and the fixed characteristics of uncertainty in responses (assumptions A1 and A2).}

{Our experiments demonstrate that valence and arousal are the two
most effective dimensions that can be used to analyze the reliability of subjects. 
Although subjects may not reach a consensus at local scales (say, an individual task) because the emotions are inherently subjective,
consensus at a global scale can still be well justified.} \\

\noindent{\textbf{Usage Scenarios:} 
We would like to articulate on the scenarios under which our method or other traditional approaches ({\it e.g.}, those described in Section~\ref{sec:baseline}) are more suitable.}

{
First, our method is not meant to replace traditional approaches that add control factors at the design stage
of the experiments, for example, recording task completion time, and testing subjects with examples annotated with gold standard labels. Those methods 
are effective at identifying extremely careless subjects. But we argue that 
the reliability of a subject is often not a matter of yes or no, but can take a continum of intermediate levels. Moreover, consensus models such as Dawid-Skene methods 
require that each task is assigned to multiple annotators.}

{
Second, our method can be integrated with other approaches so as to collect data most efficiently.   
Traditional heuristic approaches require the host to come up with a number of design questions or procedures effective for screening spammers before executing the experiments, which can be a big challenge especially for affective data.
In contrast, the consensus models 
support post analyses of collected data and have no special requirement for the experimental designs. This suggests we may use a consensus model to carry out a pilot study which then informs us how to best design the data collection procedure.}

{
Third, as a new method in the family of consensus models, our approach is unique in terms of its fundamental assumptions, and hence
should be utilized in quite different scenarios than the other models. Methods based on modeling confusion matrix are
more suitable for aggregating binary and categorical labels, while the agreement-based methods (ours included) are more suitable for 
continuous and multi-dimensional labels (or more complicated structures) 
that normally have no ground truth. The former are often evaluated quantitatively by
how accurately they estimate the true labels~\cite{sheshadri2013square}, while the latter are evaluated directly by
how effectively they identify unreliable annotators, a perspective barely touched in the existing literature.}\\

\noindent\textbf{Limitations and Future Work:} 
Despite the fact that we did not assume A1 or A2 and {approached the problem of assessing the quality of crowdsourced data form an unusual angle}, there are interesting questions left about the statistical model we employed.
\begin{itemize}
\item
{Some choices of parameters in the model are quite heuristic. The usage of our model requires pre-set values for certain parameters, {\it e.g.}, $\gamma$, but we have not found theoretically pinned-down guidelines on how to choose those parameters. As a result, it is always subjective to some extent to declare a subject spammer. The ranking of reliability of subjects seems easier to accept. Where the cutoff should be will involve some manual checking on the result or will be determined by some other factors such as the desired cost of acquiring a certain amount of data. }
\item
{Although we have made great efforts to design various measures to evaluate our method, struggling to get around the issue of lacking an objective gold standard (its very existence has been questioned), these measures have limitations in one way or the other, as discussed in Section~\ref{sec:exp}. We feel that due to the subjective nature of emotion responses to images, there is no simple and quick solution to this. The ultimate test of the method has to come from its usage in practice and a relatively long-term evaluation from the real-world.}
\item The effects of subgroup consistency, though varied from task to task,
were random effects. We constructed the model this way to stretch its applicability because
the number of responses collected per task in our empirical data was often small. 
{Some related approaches ({\it e.g.}~\cite{whitehill2009whose}) propose to estimate a difficulty/consistency parameter for each task,
but often require a relatively large number of annotators per task. }
Which kind of probabilistic assumptions is more accurate or works better calls for future exploration.
\item Only one ``major'' reliable mode was assumed at one time, and hereafter only the regularities 
conditioned on this mode are estimated. In another word, all the reliable users are assumed to behave consistently. One may ask whether there exist subgroups of reliable users who behave consistently within a group but differ across groups for reasons such as different demographic backgrounds. In our current model, if such ``minor'' reliable mode exists in a population, these subjects may be absorbed into the spammer group. Our model implicitly assumes that diversity in demography or in other aspects does not cause influential differences in emotion responses. 
{Because of this, our method in dealing with culturally sensitive data is not well justified.}
\end{itemize}
{Experimentally our method is only evaluated on one particular large data set~\cite{xin2016}. 
Evaluations on other affective data sets (when publicly available) are of interest.}

{We have focused on the post analysis of collected data. 
As a future direction, it is of interest to examine the capacity of our approach to reduce time and cost in the practice of crowdsourcing using A/B test. 
We hereby briefly discuss an online heuristic strategy to dynamically allocate tasks to more reliable subjects. 
Recall that our model has two sets of parameters:
parameter $\tau_i$ indicating the reliability of subjects and parameter $\alpha_i; \beta_i$ capturing the regularity. 
We can use the variance of distribution $\textrm{Beta}(\alpha_i, \beta_i)$ to determine how confident we are with the estimation 
of $\tau_i$. For subject $i$, if the variance of $\textrm{Beta}(\alpha_i, \beta_i)$ 
is smaller than a threshold while $\tau_i$ is below a certain percentile, 
this subject is considered {\em confidently} unreliable and he/she may be excluded from the future subject pool.}

\section{Conclusions}
In this work, we developed a probabilistic model, namely Gated Latent Beta Allocation, to analyze
the off-line consensus for crowdsourced affective data. 
Compared to the usual crowdsourcing settings, where reliable 
workers are supposed to have consensus, the consensus
analysis of affective data is more challenging because of the innate variation in emotion responses even out of true feelings.  To overcome this difficulty,  our
model estimates the reliability of subjects by exploiting the agreement relationships between their ratings at a global scale. The experiments show that the relational data based on the valence of human responses are more effective than the other emotion dimensions for identifying spammer subjects. 
By evaluating and comparing the new method with some standard methods in multiple ways, we find that the results have demonstrated clear advantages and the system seems ready for use in practice.

\section*{Acknowledgments}
This material is based upon work supported by the National Science Foundation under Grant No. 1110970. We are grateful to the reviewers and the Associate Editor for their constructive comments.

\bibliographystyle{IEEEtran}

\bibliography{emotion}

\begin{IEEEbiography}[{\includegraphics[height=1.3in, trim={30 40 20 5}, clip,keepaspectratio]{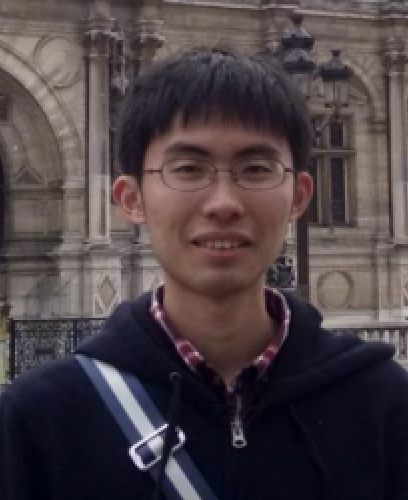}}]{Jianbo Ye} 
received his B. S. degree in Mathematics from the
University of Science and Technology of China in 2011. 
He worked as a research postgraduate at
The University of Hong Kong, from 2011 to 2012,
and a research intern at Intel Labs, China in 2013. 
He is currently a PhD candidate and Research Assistant at the 
College of Information Sciences and Technology,
The Pennsylvania State University.
His research interests include statistical modeling
and learning, numerical optimization and method, and affective image modeling.
\end{IEEEbiography}

\begin{IEEEbiography}[{\includegraphics[width=1.2in,height=1.3in, trim={0 35 20 0}, clip,keepaspectratio]{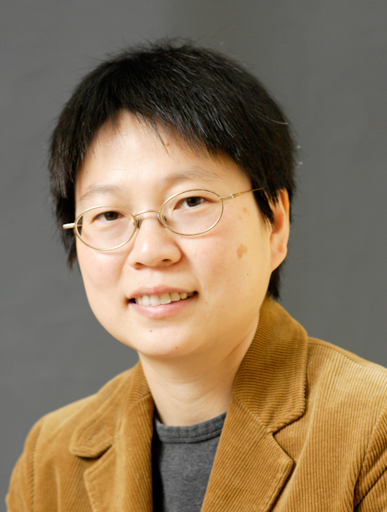}}]{Jia Li} is a Professor of Statistics at The Pennsylvania State University. She received the MS degree in Electrical Engineering, the MS degree in Statistics, and the PhD degree in Electrical Engineering, all from Stanford University. She worked as a Program Director in the Division of Mathematical Sciences at the National Science Foundation from 2011 to 2013, a Visiting Scientist at Google Labs in Pittsburgh from 2007 to 2008, a researcher at the Xerox Palo Alto Research Center from 1999 to 2000, and a Research Associate in the Computer Science Department at
Stanford University in 1999. Her research interests include statistical modeling and learning, data mining, computational biology, image processing, and image annotation and retrieval.
\end{IEEEbiography}
\vspace{-0.25in}

\begin{IEEEbiography}[{\includegraphics[width=1.2in,height=1.3in, trim={2 8 2 8},clip,keepaspectratio]{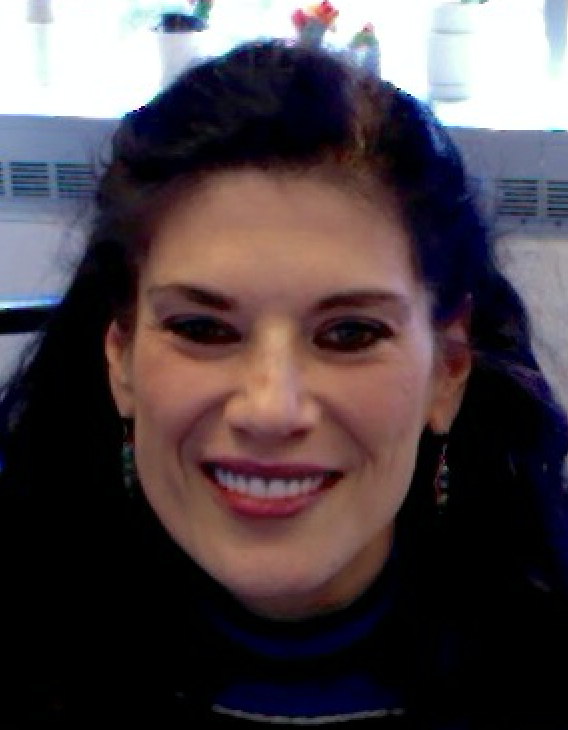}}]{Michelle G. Newman}  is a Professor of Psychology and Psychiatry, and the Director of the Center for the Treatment of Anxiety and Depression at The Pennsylvania State University.  She received her PhD from the State University of New York at Stony Brook in 1992 and completed a post-doctoral Fellowship at Stanford University School of Medicine in 1994. She has conducted psychotherapy outcome studies for generalized anxiety disorder, social phobia, and panic disorder, and has done basic emotion and experimental work related to these disorders. She is currently an editor for Behavior Therapy and is on the editorial boards of Psychotherapy Research, Cognitive Therapy and Research, and American Journal of Health Behavior. She is the recipient of the American Psychological Association (APA) Division 12 Turner Award for distinguished contribution to clinical research, and the APA Society of Psychotherapy (Division 29): Distinguished Publication of Psychotherapy Research Award. She is a Fellow of APA Divisions 12 and 29 and of the American Association for Behavioral and Cognitive Therapies.
\end{IEEEbiography}
\vspace{-0.25in}

\begin{IEEEbiography}[{\includegraphics[width=1.2in,height=1.3in, trim={10 0 10 5}, clip,keepaspectratio]{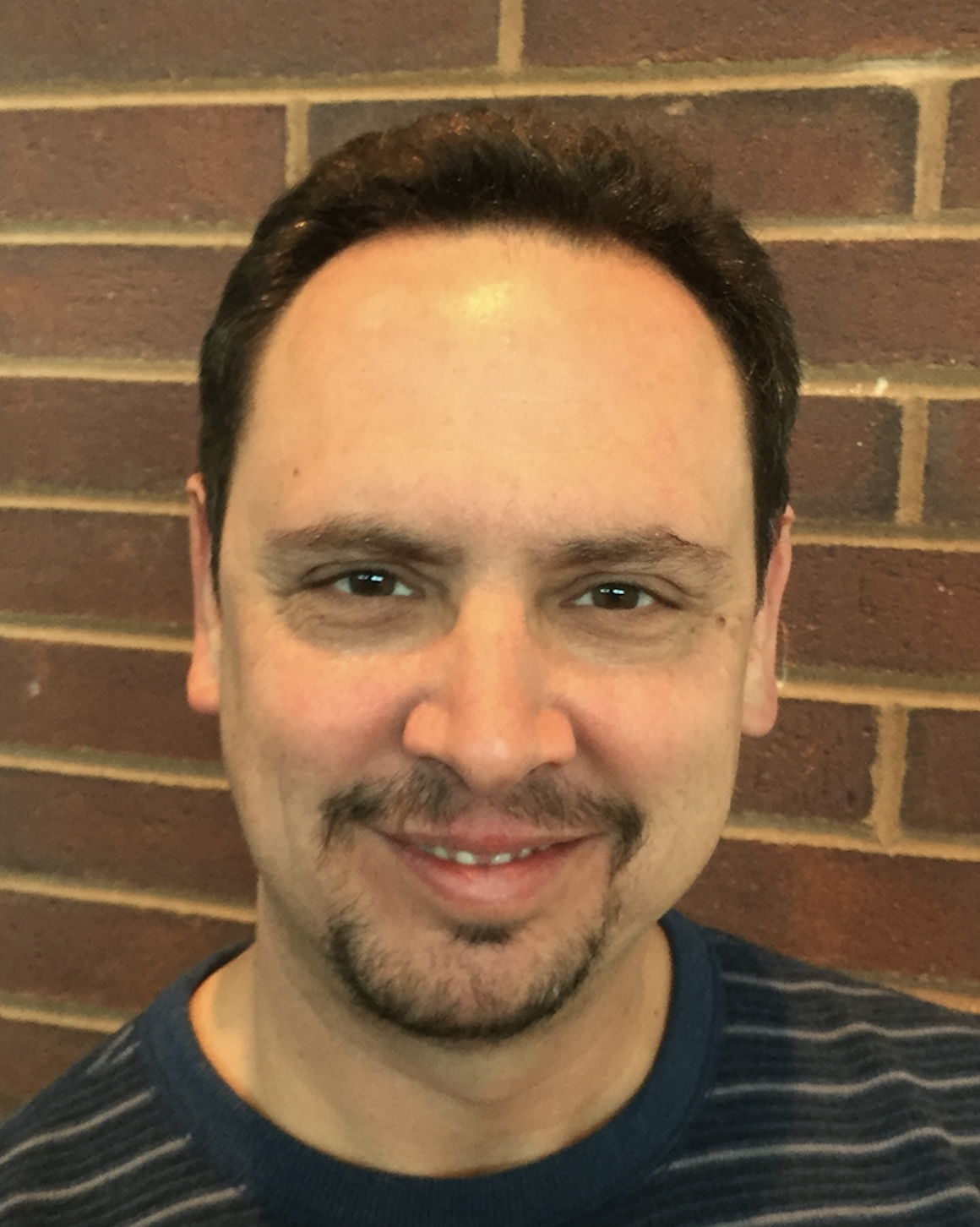}}]{Reginald B. Adams, Jr.} is an Associate Professor of Psychology at The Pennsylvania State University. He received his PhD from Dartmouth College in 2002. He is interested in how we extract social and emotional meaning from nonverbal cues, particularly via the face. His work addresses how multiple social messages ({\it e.g.}, emotion, gender, race, age, etc.) combine across multiple modalities and interact to form unified representations that guide our impressions of and responses to others. Although his questions are social psychological in origin, his research draws upon visual cognition and affective neuroscience to address social perception at the functional and neuroanatomical levels. Before joining Penn State, he was awarded a National Research Service Award (NRSA) from the National Institute of Mental Health to train as a postdoctoral fellow at Harvard and Tufts Universities. His continuing research efforts have been funded through NSF, NIA and NIMH (NIH). 
\end{IEEEbiography}
\vspace{-0.25in}

\begin{IEEEbiography}[{\includegraphics[width=1.2in,height=1.3in, clip,keepaspectratio,trim={0 0 0 0}]{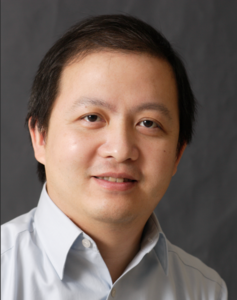}}]{James Z. Wang} is a Professor of Information Sciences and Technology at The Pennsylvania State University. He received the bachelor's degree in mathematics and computer science {\it summa cum laude} from the University of Minnesota, and the MS degree in mathematics, the MS degree in computer science and the PhD degree in medical information sciences, all from Stanford University. His research interests include computational aesthetics and emotions, automatic image tagging, image retrieval, and computerized analysis of paintings. He was a visiting professor at the Robotics Institute at Carnegie Mellon University (2007-2008), a lead special section guest editor of the IEEE Transactions on Pattern Analysis and Machine Intelligence (2008), and a program manager at the Office of the Director of the National Science Foundation (2011-2012). He was a recipient of a National Science Foundation Career award (2004).
\end{IEEEbiography}


\end{document}